\documentclass[journal]{IEEEtran}
\usepackage[T1]{fontenc}
\usepackage[utf8]{inputenc}
\usepackage{graphicx,amsmath,amssymb,booktabs,multirow,array,makecell,cite}
\usepackage{tabularx}
\usepackage[hidelinks]{hyperref}
\newcommand{\Coarse}{\textit{coarse}}
\newcommand{\FullDesc}{\textit{full}}
\newcommand{\CoarsePath}{\Coarse{}$\rightarrow$path}
\newcommand{\FullPath}{\FullDesc{}$\rightarrow$path}
\newcommand{\CoarseFull}{\Coarse{}$\rightarrow$\FullDesc{}}
\newcommand{\auditcell}[2]{#2}
\usepackage{xcolor}
\usepackage{hyperref}

\begin{document}
\title{RemoteVLPG: How to Obtain a Specific VLM for Language-Guided Path Generation on Remote Sensing Images?}
\author{Junzhe Dang, Mengmeng Zhang, Yuxiang Zhang, and Wei Li,~\IEEEmembership{Senior Member,~IEEE}
    \thanks{Junzhe Dang, Mengmeng Zhang, and Wei Li are with the School of Information and
        Electronics, Beijing Institute of Technology,  and
        the National Key Laboratory of Science and Technology on Space-Born
        Intelligent Information Processing, Beijing
        100081, China, and also with  Beijing Institute of Technology, Zhuhai, Guangdong 519088, China. (e-mail: dangjunzhe@bit.edu.cn; mengmengzhang@bit.edu.cn; liwei089@ieee.org). Corresponding author: Mengmeng Zhang.}
    \thanks{Yuxiang Zhang is with the Jockey Club STEM Laboratory of Quantitative Remote Sensing, Department of Geography, the University of Hong Kong, Hong Kong, China. (e-mail: yxzhang7@hku.hk).}}

\maketitle

\begin{abstract}
    We study how to obtain a task-specific Vision-Language Model (VLM) for language-guided path generation between start and end points on remote sensing images. We explore this question from three aspects. First, we implement three widely used architectural approaches and compare their performance. Second, we design a scalable method for constructing pre-training data and build an effective pre-training dataset with a degree of diversity. Methods trained on this dataset achieve promising zero-shot performance. Third, we construct three benchmarks that support stratified evaluation and design three metrics for path generation. We investigate this problem through extensive experiments. The results show that the zero-shot performance of methods trained on the constructed pre-training dataset can exceed that of some advanced commercial VLMs. Code and Dataset can be available at \textcolor{blue}{\textbf{\small\url{https://github.com/jzdang/RemoteVLPG}}}
\end{abstract}


\section{Introduction}

General-purpose language models have gradually developed broader capabilities. However, fine-grained visual-language capabilities approaching annotation-level accuracy require models with large amounts of training data and parameters, such as Claude Fable 5.2~\cite{anthropic2026claudefable52} or Gemini 4 Carbon~\cite{googledeepmind2026gemini4carbon}. These models are expensive and unaffordable for many users, and waiting for API responses takes time. Therefore, research on task-specific visual-language models remains active~\cite{lai2024lisa,bao2024cores,quenum2026lisat,qian2025reasoning,kirillov2023segment,zhang2024psalm,li2025segearth,xin2026segearth}.

Research on specific visual-language tasks mainly combines language with mainstream visual tasks. Many works studies produce textual or symbolic answers, including neuro-symbolic reasoning~\cite{wei2026geoint,yang2026nesy} and visual question answering (VQA)~\cite{li2023blip,liu2023visual}. For visual perception, research mainly introduces language guidance into mainstream visual tasks, such as referring segmentation~\cite{lai2024lisa,bao2024cores,quenum2026lisat,qian2025reasoning}, reasoning segmentation~\cite{li2025segearth,xin2026segearth}, and language-guided object detection~\cite{du2024lami,jain2022bottom,wang2025ov}. These methods still address mainstream visual tasks, while multimodal language guidance enables interaction between human and machine: users can specify the target to be inferred through language. Tasks beyond mainstream visual perception that become possible through multimodal technology remain less explored.

Benefiting from advances in multimodal technology, we investigate path generation, a task that has received relatively little attention in mainstream visual research. Studying path generation with a single modality is difficult because the desired path between two points is hard to specify. Specifically, we study how to generate a desired path from a top-down remote sensing image under language guidance. Users can communicate their requirements through language and obtain corresponding path coordinates. This research has value in two ways. First, top-down remote sensing images reveal many ground objects. Given start and end coordinates and a human description of the desired path, it may be possible to generate path coordinates that guide the movement of ground objects. Second, a multimodal path generation model can provide path priors to other task-specific models and help train them.

Our goal is to train a task-specific VLM for language-guided path generation. We aim for a model with a few billion parameters so that it can be deployed more easily as edge computing capabilities improve. Avoiding network requests to high-performance but expensive general-purpose VLM APIs may also improve responsiveness and reduce cost. We therefore consider the following questions.

1) \textbf{What architecture should the model use?} This is the algorithm design question.

2) \textbf{How should model performance be evaluated?} This involves constructing two or three dedicated evaluation datasets and designing appropriate metrics for path quality.

3) \textbf{Does a VLM dedicated to language-guided path generation from top-down remote sensing images require task-specific pre-training?} If so, a pre-training dataset is needed. This question concerns both the need for pre-training and the construction of its data.

Investigating all three questions is essential to this study. We address them individually and describe our solutions in detail in the following sections. We first outline our answers.

\begin{figure*}[!tp]
    \centering
    \includegraphics[width=\textwidth]{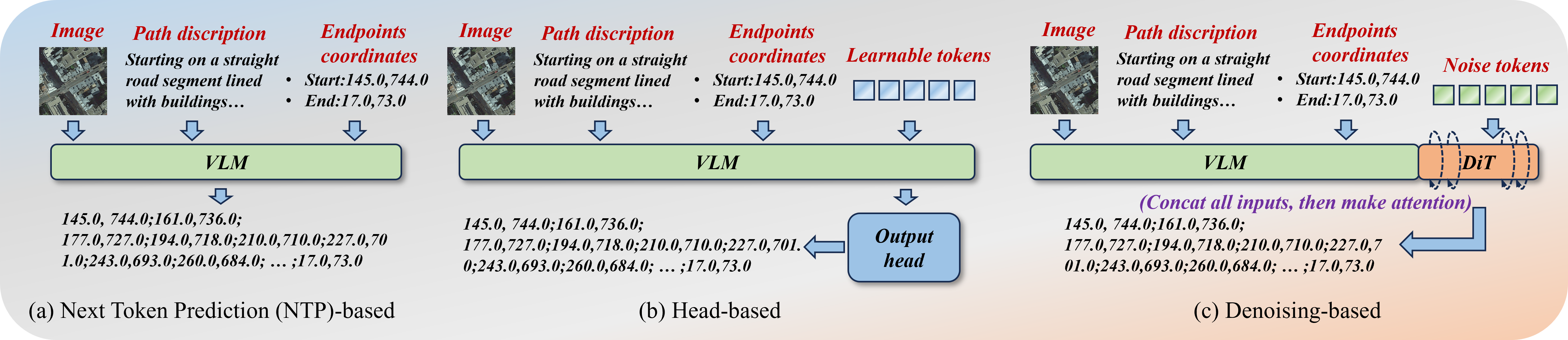}
    \caption{Architectures of the three methods: (a) NTP-based, (b) Head-based, and (c) Denoising-based.}\label{fig:1}
\end{figure*}

1) Algorithm design is important because model architecture and training trajectory determine how a method converges and achieve its final performance. \textbf{We consider three widely used VLM architectures}, as shown in Fig.~\ref{fig:1}.

The first is autoregressive generation, also called a Next-Token Prediction (NTP)-based method, as shown in Fig.~\ref{fig:1}(a). GeoChat~\cite{kuckreja2024geochat} and Falcon~\cite{yao2025falcon} use this achitecture. The image, language description, and start and end coordinates are all provided to the VLM, which autoregressively generates ordered point coordinates from the start to the end of the path. Its training procedure is consistent with that of most VLMs and can make good use of pre-trained knowledge. However, its inference output may violate the required format.

The second is a head-based method, as shown in Fig.~\ref{fig:1}(b). A fixed number of tokens are usually taken from the VLM output and passed to a decoder, whose output is compared with ground truth to compute the loss. We design a simple path decoder with several Transformer~\cite{vaswani2017attention} blocks. Mean squared error (MSE) between its predicted path and the ground-truth coordinates is used to optimize the whole VLM. Representative methods include LISA~\cite{lai2024lisa}, CoReS~\cite{bao2024cores}, LISAT~\cite{quenum2026lisat}, and READ~\cite{qian2025reasoning}, which use SAM~\cite{kirillov2023segment} as an output head, and PSALM~\cite{zhang2024psalm}, SegEarth-R1~\cite{li2025segearth}, SegEarth-R2~\cite{xin2026segearth}, and UniGeoSeg~\cite{ni2026unigeoseg}, which use Mask2Former~\cite{cheng2022masked}. This architecture follows the encoder--decoder design common in visual tasks. The decoder output is directly compared with ground truth, making the loss and training procedure intuitive and allowing improvements based on experience with visual tasks.

The third is a denoising-based method, as shown in Fig.~\ref{fig:1}(c). Denoising mode is widely used in image generation, where many methods are based on diffusion~\cite{ho2020denoising} or flow matching~\cite{lipman2022flow}. Inspired by the action experts in embodied vision-language-action (VLA) models such as $\pi_0$~\cite{black2024pi_0}, $\pi_{0.5}$~\cite{intelligence2025pi_}, and Qwen-VLA~\cite{wang2026qwen}, we use a denoising expert to denoise paths. As in these VLA models, the path expert uses a diffusion transformer (DiT) architecture and is trained through flow matching. During denoising, it obtains visual and language conditioning features from the VLM through attention. This architecture explores the value of a successful VLA paradigm for our task.

The detailed designs of these three methods are presented in the methods section.

2) For performance evaluation, \textbf{we first construct two evaluation benchmarks from existing datasets and a third from images of Shenyang and Tokyo collected using Google Earth Studio}\footnote{Google Earth Studio: \url{https://www.google.com/earth/studio/}.}. We introduce stratified evaluation into each benchmark. \textbf{We also propose three metrics for evaluating path generation}. The benchmark construction and metric definitions are detailed later.

3) To investigate the need for task-specific pre-training and the construction of its data, \textbf{we first build a relatively large pre-training dataset for a VLM dedicated to language-guided path generation}. We pre-train all three architectures on this dataset and evaluate their zero-shot performance on the test splits of the three benchmarks. We also compare fine-tuning on the training splits of two benchmarks with and without task-specific pre-training, followed by evaluation on their test splits, to examine whether task-specific pre-training is necessary.

The main contributions of this work are summarized as follows. We investigate language-guided path generation on remote sensing images, a task that has received relatively little attention, and conduct a comprehensive study of this problem from three aspects.
\begin{itemize}
     \item We investigate three widely used VLM architectural approaches for language-guided path generation, i.e., Next-Token Prediction (NTP)-based, head-based, and denoising-based methods. We implement these three approaches and systematically compare their architectural differences and performance.
    
     \item We construct a large-scale, task-specific pre-training dataset, RemotePath-175k, and evaluate the zero-shot performance of all three approaches after pre-training on this dataset. Meanwhile, we develop a scalable image–path alignment method, making it easier to further expand the dataset.
    
     \item We construct three benchmarks for evaluating language-guided path generation, i.e., LPBench, PPBench, and ToSh-Data. All three benchmarks support stratified evaluation based on path pixel length and directional trend count. In addition, we propose three metrics, i.e., Normalized Average Displacement Error (NADE), Normalized Chamfer Distance (NCD), and Path Length Error (PLE), to evaluate the quality of generated paths.
\end{itemize}
\section{Related Work}


\subsection{Methods based on Output Head}
Head-based VLMs mainly address language-guided mainstream visual tasks. Segmentation methods such as LISA, LISAT, READ, CoReS, PSALM, and SegEarth-R1 use SAM or Mask2Former as a decoder. UniChange~\cite{zhang2026unichange} also uses SAM for change detection. Object detection methods such as LaMI-DETR~\cite{du2024lami}, BUTD-DETR~\cite{jain2022bottom}, and OV-DQUO~\cite{wang2025ov} use DETR~\cite{carion2020end} to decode detection results. This achitecture can directly transfer advanced decoders to language-guided visual tasks. Its architecture is intuitive, and losses from the corresponding visual tasks can directly optimize the model.

\subsection{Diffusion Transformer (DiT)}
DiT~\cite{peebles2023scalable} replaces the U-Net~\cite{ronneberger2015u} in the original diffusion design with a Transformer. As Transformers have become a mainstream architecture, attention allows DiT to connect readily to other Transformer-based models, including Large Language Models (LLMs) and VLMs. Compared with U-Net-based image denoising, the Transformer architecture also extends the shapes of objects that can be denoised, enabling generation of objects with different shapes. These advantages support connections between models and generation of diverse objects, greatly expanding the use of generative models. Inspired by the success of DiT in VLA models~\cite{black2024pi_0,intelligence2025pi_,wang2026qwen}, we construct path denoising expert based on DiT as one of our three architectural approaches.

\subsection{Methods based on Autoregressive Generation}
Autoregressive VLMs mainly address VQA and language-guided mainstream visual tasks. VistaLLM~\cite{pramanick2024jack} autoregressively outputs boundary coordinates of segmentation masks for language-specified objects. GeoChat also directly outputs object coordinates as numerical text. This approach can make substantial use of the numerical knowledge in VLM pre-training. Falcon uses autoregressive generation but adds new vocabulary IDs for discrete coordinates quantized to the image dimensions, instead of producing numerical text. When decoding features into IDs, it samples only from those representing discrete coordinates. This restriction avoids the invalid outputs that may occur with direct text prediction in GeoChat. However, each training sample contains only a few coordinates. If the dataset is not large enough, not all discrete coordinates in the image can be sufficiently trained. An insufficiently trained ID has no corresponding knowledge in the model's original pre-training and may therefore cause test errors. Considering these trade-offs, we follow GeoChat and construct our autoregressive path generation method by directly outputting coordinates as numerical text.

\section{Three VLM Architectures Based on PaliGemma 2 3B}
\begin{figure}[!tp]
    \centering
    \includegraphics[width=\columnwidth]{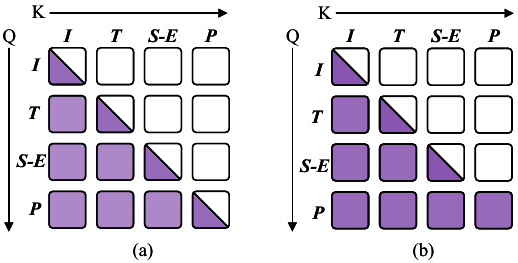}
    \caption{(a) Attention masks of the NTP-based and head-based methods. (b) Attention mask of the denoising-based method. Here, I, T, S-E, and P denote image, text, start and end coordinates, and path (ground truth, learnable tokens, or noise), respectively.}\label{fig:10}
\end{figure}
\subsection{Head-based Method}
As shown in Fig.~\ref{fig:1}(b), the image, text description, start and end points, and learnable tokens are concatenated in this order and input to the VLM. These tokens follow strictly causal attention. Image tokens cannot attend to later inputs and use unidirectional attention within their block. Text tokens attend to all image tokens and use unidirectional attention internally. The start and end coordinates and their prompts also use unidirectional attention within their block and can attend to the image and text tokens. Learnable tokens use unidirectional attention internally and can attend to all other preceding blocks. This globally causal attention makes good use of VLM pre-trained knowledge, as shown in Fig.~\ref{fig:10}(a).

During path generation, autoregressive generation is disabled. The trained VLM acts as a multimodal encoder. Outputs at the positions of the learnable tokens are extracted and passed to the output head, which decodes the path coordinates.

The output head has two Transformer layers, whose embedding dimension is 2304, matching PaliGemma 2 3B~\cite{steiner2024paligemma}. Each layer has eight attention heads, and the intermediate dimension of its multilayer perceptron (MLP) is 4096.

We use a simple MSE loss. Each path is sampled at equal arc-length intervals into 20 points, including the start and end points. During training, each point's coordinates are normalized to $[-1,1]$. The start and end predictions also contribute to the loss and backpropagation. At inference, we replace the start and end predicted points with the true start and end coordinates before evaluation.

\subsection{NTP-based Method}
As shown in Fig.~\ref{fig:1}(a), the image, language description, start and end coordinates, and ground-truth path coordinates are input to the language model in this order. Training uses teacher forcing and a cross-entropy loss for next-token prediction. Attention is causal between blocks: tokens in a later block can attend to all tokens in preceding blocks. It is also causal within each block, where a token can attend only to earlier tokens. This attention mask is designed to make full use of the VLM's pre-trained knowledge, as shown in Fig.~\ref{fig:10}(a).

For autoregressive training, target coordinates are first normalized to $[-1,1]$, as in the head-based and denoising-based methods, and then rescaled to $[0,100]$ as the training targets, which follows GeoChat.

At inference, ground-truth path coordinates are omitted. Student forcing are adopted, where the model's own outputs are fed back step by step, autoregressively generates the complete ordered path coordinates, including the start and end points.

\subsection{Denoising-based Method}
As shown in Fig.~\ref{fig:1}(c), the path expert performs joint attention with the VLM during training. It has four Transformer blocks that attend to the hidden states of layers 7, 13, 20, and 26 of PaliGemma 2 3B in sequence. Because the VLM hidden size is 2304 and the path expert hidden size is 512, additional linear layers map the VLM features from 2304 to 512 dimensions.

The VLM receives the image, language, and start and end coordinates, whereas the path expert receives path noise. Inside the VLM, the image, language, and coordinate blocks follow causal attention in that order, with causal attention also within each block, to make full use of pre-trained knowledge. Path noise tokens can attend to all VLM tokens. Within the path expert, they use bidirectional attention so that each path point can obtain information from points both before and after it, as shown in Fig.~\ref{fig:10}(b).

The path expert contains four Transformer blocks and uses Grouped-Query Attention (GQA)~\cite{ainslie2023gqa}. Queries have eight heads with 64 dimensions per head, whereas keys and values have four heads with 64 dimensions per head. The MLP intermediate dimension is 2048.

Flow matching is adopted when training. At inference, the ordinary differential equation (ODE) solver uses 20 steps. As in the head-based method, ground-truth path coordinates are rescaled to $[-1,1]$ before training.

\section{Construction of Pre-training Data}

\begin{figure*}[!t]
    \centering
    \includegraphics[width=\textwidth]{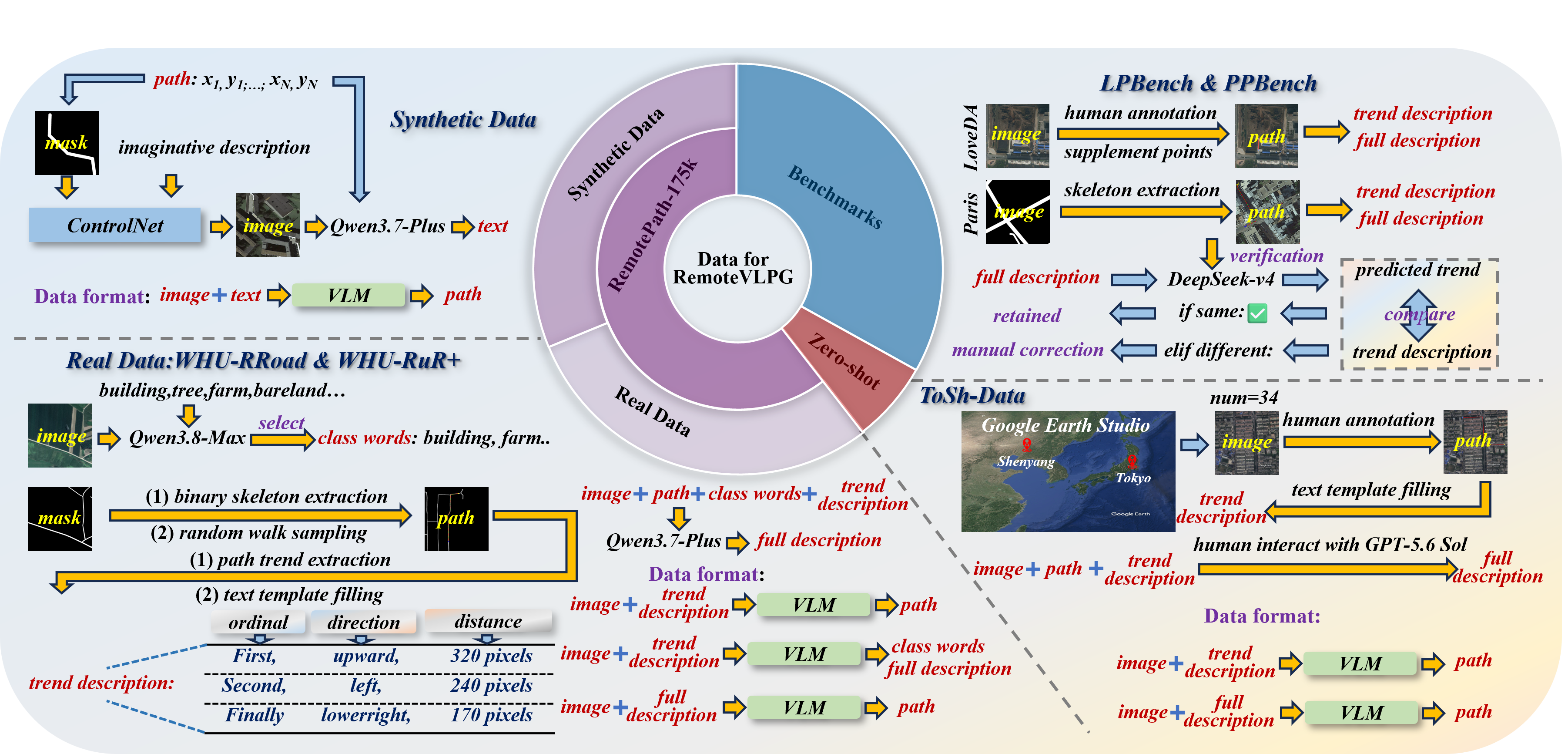}
    \caption{Construction workflow for the pre-training dataset and the three benchmarks: LPBench, PPBench, and ToSh-Data.}\label{fig:2}
\end{figure*}

We construct RemotePath-175k to pre-train the three architectures based on PaliGemma 2 3B, which contains synthetic and real data, as shown in Fig.~\ref{fig:2}.

\begin{figure*}[!t]
    \centering
    \includegraphics[width=\textwidth]{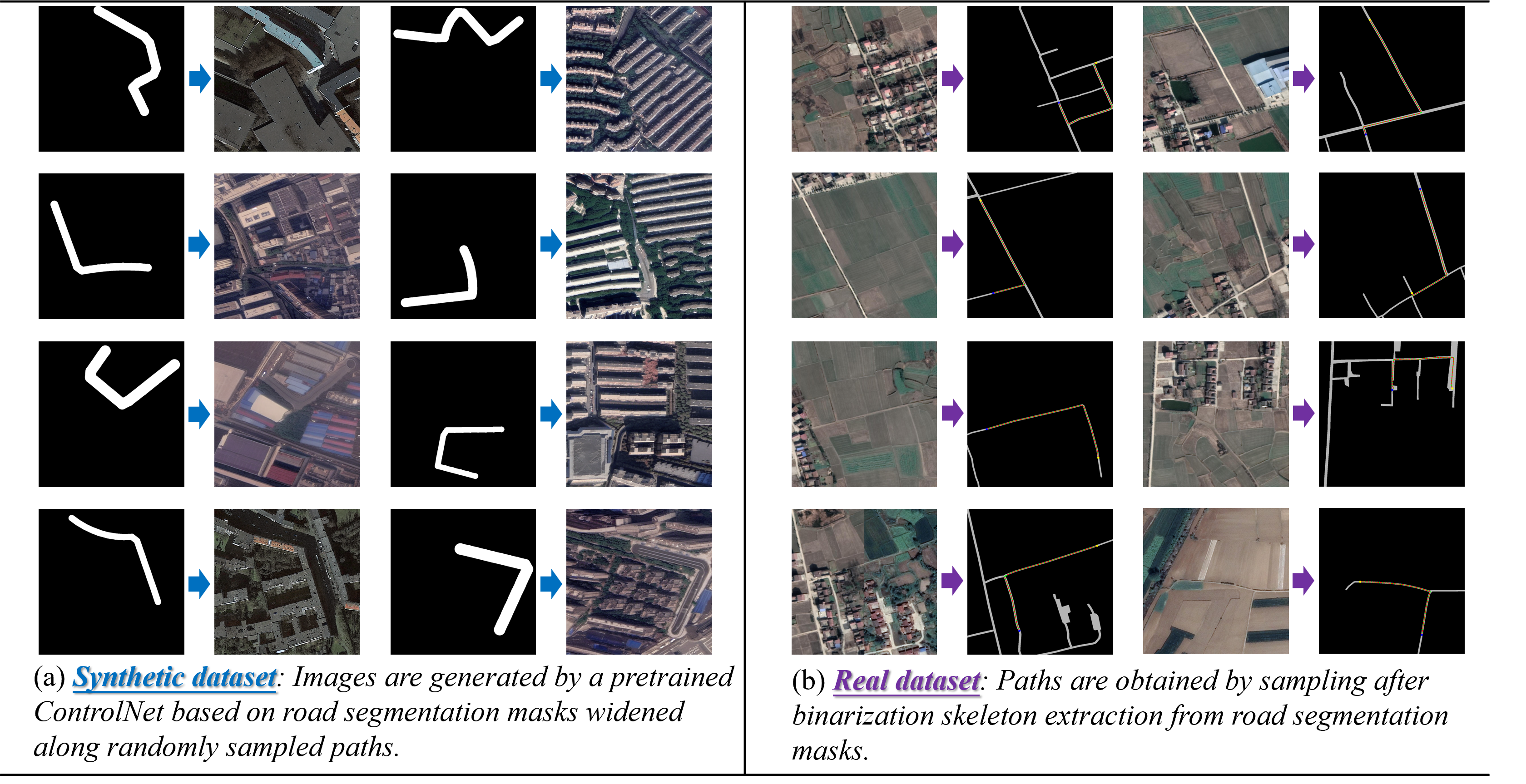}
    \caption{Image--path alignment in the synthetic and real pre-training datasets. (a) The trained ControlNet generates images from segmentation masks obtained by widening randomly sampled paths. (b) Binary morphological skeletonization extracts paths from real road segmentation masks.}\label{fig:3}
\end{figure*}

For the synthetic component, we collect road segmentation data from several datasets to train ControlNet~\cite{zhang2023adding}. To generate data, we randomly generate paths according to rules such as the number of turns, widen them into road masks, and input the masks to the trained ControlNet to obtain synthetic images, as shown in Fig.~\ref{fig:3}(a). This naturally matches image content to path shape and removes the need to manually annotate paths on images. The trained ControlNet can generate an unlimited number of path-matched images, allowing many synthetic samples to be obtained. We then use commercial VLMs to generate corresponding path descriptions. For the real component, we collect two additional road segmentation datasets, extract road skeletons, and sample along them according to requirements such as turn count and length to obtain paths.

\subsection{Synthetic Data}
We use the Berlin and Chicago road segmentation datasets~\cite{kaiser2017learning}, ISPRS Potsdam dataset\footnote{ISPRS 2D Semantic Labeling Benchmark - Potsdam Dataset, provided by the German Society for Photogrammetry, Remote Sensing and Geoinformation (DGPF).}, CHN6-CUG Roads Dataset~\cite{zhu2021global}, DeepGlobe Road Extraction Challenge dataset~\cite{demir2018deepglobe}, and Ottawa-dataset~\cite{liu2018roadnet}. Images and road masks are jointly cropped and resized to $512\times512$ to train ControlNet. The model loads Stable Diffusion 1.5~\cite{rombach2022high} weights. It receives noise image, a segmentation mask, and a text description and outputs the corresponding image. Training images and masks come from these six datasets. Qwen3-VL-Plus~\cite{bai2025qwen3} generates each training description from the road shapes in the image, describing roads and the scene in no more than 25 words.

Once trained on these real images, text descriptions, and road masks, ControlNet is used to generate synthetic data. At inference for data creation, it requires a road segmentation mask and a text description to generate the corresponding image.

\subsubsection{Random path generation}
We randomly sample path coordinates. A start point is randomly placed within an image, and the number of subpaths is specified. Each subpath can be a straight line or an arc, with straight lines sampled more frequently. A path contains at most four or five subpaths, and two or three subpaths are assigned higher sampling probabilities. Thus, sampled paths contain some turns without excessive turning trends. We also constrain the pixel length of each subpath and the total pixel length of the path.

These constraints keep the sampled path lengths and turn counts within reasonable ranges. Direction checks, subpath intersection detection, and path self-intersection detection further constrain the path shapes.

We use each generated path as a road centerline and widen it into a road segmentation mask. Road width is independently sampled for each path within a specified pixel-width range, keeping widths reasonable while allowing variation. We also widen a selected subset into 4000 narrow-road masks to increase the number of narrow-path samples in pre-training.

\subsubsection{Text input for ControlNet inference}
We prompt Qwen3.6-Flash~\cite{qwen36_35b_a3b} to imagine a description of the path shape and scene for each path. This description and its widened road mask are fed to well-trained ControlNet, generating an image aligned with the road mask and hence with the path.

We generate 24,000 images in this way. Training the three architectures requires images and path descriptions. For the synthetic component, we do not use the imagined Qwen3.6-Flash descriptions supplied to ControlNet. Instead, Qwen3.7-Plus~\cite{qwen37plus} generates new descriptions from the resulting images and paths. Each synthetic sample takes a full path description (\FullDesc{}) together with the image and start and end coordinates as input and outputs ordered path coordinates (\FullPath{}). The synthetic component contains 24,000 such samples.

\subsection{Real Data}
We use WHU-RRoad~\cite{9824c6ce552244e087dd5d4f7ad23883} and WHU-RuR+~\cite{wang2025whu}. Binary morphological skeletonization from the established scikit-image library~\cite{van2014scikit} reduces each road segmentation mask to a one-pixel-wide skeleton. Sampling from a point along the skeleton yields a complete path, as shown in Fig.~\ref{fig:3}(b). This directly converts road masks into path coordinates and avoids costly manual annotation. To obtain language descriptions for training, we use a three-stage procedure to produce a coarse trend description (\Coarse{}) and a \FullDesc{} description.

\subsubsection{Obtaining \Coarse{} text}
We use a fixed template to construct \Coarse{} text . We calculate the angle between adjacent path vectors and identify a trend change when it exceeds $20^\circ$, filtering out consecutive change points that are too close. Each segment direction is quantized to one of eight trends: left, right, upward, downward, upper left, lower left, upper right, and lower right. We calculate the pixel length of each trend segment and fill its direction and length into the template. Segments are connected with words such as First, Next, Then, and Finally. One resulting description is: ``From the starting point, First,the road extends approximately 757 pixels toward the upward direction. Finally, the road extends approximately 115 pixels toward the right direction.''

\subsubsection{Obtaining class nouns}
We manually inspect images in WHU-RRoad and WHU-RuR+ and identify seven classes: ``building, field path, wide road, farmland, bare land, water, forest''. The advanced multimodal model Qwen3.8-Max~\cite{qwen38} receives these candidate nouns and an image, and outputs the class nouns present in that image. This provides class information for every image.

\subsubsection{Obtaining \FullDesc{} text}
Qwen3.7-Plus receives the image, class nouns, \Coarse{} text, and path coordinates and generates a \FullDesc{} description of scene changes from the start to the end. Compared with \Coarse{}, it additionally describes changes in the scene along both sides of the path and thus contains more information.

We use these three types of text to construct three training tasks. In addition to the image and start and end coordinates, the first takes \Coarse{} as input and outputs ordered path coordinates (\CoarsePath{}). The second is \FullPath{}, as defined above. The third takes \Coarse{} as input and outputs a \FullDesc{} description (\CoarseFull{}). All three tasks are conditioned on the image and start and end coordinates, and the arrows describe only the relationship between language input and language output. In \CoarseFull{}, the model first outputs the classes in the image and then the \FullDesc{} text. This introduces a simple template-based Chain-of-Thought (CoT)~\cite{wei2022chain} reasoning process, for example: ``Think: I recognize the following classes in the image:building,field path,farmland,water,forest. Answers:Starting from the south,the path moves upward along the edge of a large water body,with buildings and farmland visible to the left.As it continues upward,the scenery remains consistent with water on the right and mixed land use on the left. Finally,the path turns right,reaching the endpoint near the water's edge,surrounded by open fields and sparse vegetation.'' The \CoarsePath{} and \FullPath{} samples train all parts of each architecture, while the \CoarseFull{} samples train only the VLM component and do not train the output head or path denoising expert.

We obtain 32,718 samples covering the three tasks from WHU-RRoad and 32,841 from WHU-RuR+, giving 65,559 real-data samples. Together with the synthetic component, there are 89,559 samples. We also add all 85,813 training VQA samples from VRSBench~\cite{li2024vrsbench}. Their answers are very short, and they are included to retain the VLM components' basic language understanding. Table~\ref{tab:1} compares RemotePath-175k with existing visual-language pre-training datasets.

\begin{table*}[t]
\centering\caption{Comparison of the constructed pre-training dataset and benchmarks with existing datasets in tasks and text length. Our datasets focus on path generation and provide longer text.}\label{tab:1}
\small\renewcommand{\arraystretch}{1.12}\setlength{\tabcolsep}{3pt}
\begin{tabular*}{\textwidth}{@{\extracolsep{\fill}}>{\raggedright\arraybackslash}m{2.8cm}>{\raggedright\arraybackslash}m{1.15cm}>{\raggedright\arraybackslash}m{3.6cm}>{\raggedright\arraybackslash}m{1.8cm}>{\raggedright\arraybackslash}m{5.6cm}@{}}
\toprule
Dataset & Samples & Tasks & Stratified evaluation & Average text length \\
\midrule
\multicolumn{5}{c}{\textbf{\auditcell{1!A2}{Datasets for pre-training\textsuperscript{a}}}} \\
\midrule
\auditcell{1!A3}{RemoteSAM~\cite{yao2025remotesam}} & \auditcell{1!B3}{270k} & \auditcell{1!C3}{multi-modal segmentation} & \auditcell{1!D3}{---} & \auditcell{1!E3}{6.9 words} \\
\auditcell{1!A4}{PreGRES~\cite{quenum2026lisat}} & \auditcell{1!B4}{1204k} & \auditcell{1!C4}{VQA} & \auditcell{1!D4}{---} & \auditcell{1!E4}{11.12 words} \\
\auditcell{1!A5}{\textbf{RemotePath-175k}} & \auditcell{1!B5}{\textbf{175k}} & \auditcell{1!C5}{\textbf{Path generation \& Path CoT VQA \& VQA}} & \auditcell{1!D5}{---} & \auditcell{1!E5}{\textbf{5.79 words for class nouns; 31.73 words for \Coarse{}; 44.80 words for \FullDesc{}.}} \\
\midrule
\multicolumn{5}{c}{\textbf{\auditcell{1!A6}{Benchmarks\textsuperscript{b}}}} \\
\midrule
\auditcell{1!A7}{RefSegRS~\cite{yuan2024rrsis}} & \auditcell{1!B7}{4.4k} & \auditcell{1!C7}{multi-modal segmentation} & \auditcell{1!D7}{No} & \auditcell{1!E7}{3.09 words} \\
\auditcell{1!A8}{EarthReason~\cite{li2025segearth}} & \auditcell{1!B8}{14k} & \auditcell{1!C8}{multi-modal segmentation} & \auditcell{1!D8}{No} & \auditcell{1!E8}{20.89 words} \\
\auditcell{1!A9}{GRES~\cite{quenum2026lisat}} & \auditcell{1!B9}{27.6k} & \auditcell{1!C9}{multi-modal segmentation} & \auditcell{1!D9}{No} & \auditcell{1!E9}{15.04 words} \\
\auditcell{1!A10}{\textbf{LPBench}} & \auditcell{1!B10}{\textbf{3k}} & \auditcell{1!C10}{\textbf{Path generation \& Path CoT VQA}} & \auditcell{1!D10}{\textbf{Yes}} & \auditcell{1!E10}{\textbf{9.25 words for class nouns; 30.23 words for \Coarse{}; 65.62 words for \FullDesc{}}} \\
\auditcell{1!A11}{\textbf{PPBench}} & \auditcell{1!B11}{\textbf{2.2k}} & \auditcell{1!C11}{\textbf{Path generation \& Path CoT VQA}} & \auditcell{1!D11}{\textbf{Yes}} & \auditcell{1!E11}{\textbf{9.50 words for class nouns; 30.05 words for \Coarse{}; 67.01 words for \FullDesc{}}} \\
\auditcell{1!A12}{\textbf{ToSh-Data}} & \auditcell{1!B12}{\textbf{34}} & \auditcell{1!C12}{\textbf{Path generation}} & \auditcell{1!D12}{\textbf{Yes}} & \auditcell{1!E12}{\textbf{33.82 words for \Coarse{}; 67.21 words for \FullDesc{}}} \\
\bottomrule\end{tabular*}
\par\vspace{3pt}
\begin{minipage}{\textwidth}
\scriptsize\raggedright
\noindent\textsuperscript{a} The average text length for RemoteSAM is taken from UniGeoSeg~\cite{ni2026unigeoseg}. The sample count for PreGRES refers to the number of question--answer pairs reported in the original paper.\par
\noindent\textsuperscript{b} The average text lengths and sample counts for RefSegRS and EarthReason are taken from UniGeoSeg~\cite{ni2026unigeoseg}. For LPBench, PPBench, and ToSh-Data, the sample counts refer to the numbers of paths rather than the totals across the three sample types, to remain consistent with Table~\ref{tab:2}. The sample count for GRES refers to the number of annotations reported in the original paper.\par
\end{minipage}\par\vspace{-9pt}
\end{table*}

\subsection{General Knowledge Data}
We select subsets of RemotePath-175k to construct two General knowledge (GK) datasets. These are mixed with benchmark training samples during fine-tuning to introduce additional general knowledge, improve its retention, and prevent overfitting on small datasets. The two versions are General knowledge dataset v1 (GK-v1) and General knowledge dataset v2 (GK-v2). GK-v2 selects 3000 images each from the synthetic and real components and retains only \FullPath{}, giving 6000 samples. GK-v1 selects 3000 synthetic images and 7000 real images, retaining all three tasks for the real component, giving 24,000 samples.

\section{Performance Evaluation}

As discussed above, evaluating our three architectures and the selected commercial models requires two or three benchmarks and suitable metrics. We first describe benchmark construction.

\subsection{Benchmark Construction}

We construct benchmarks from two existing datasets, LoveDA~\cite{wang2021loveda} and Paris~\cite{kaiser2017learning}. We also collect 34 images of Shenyang, China, and Tokyo, Japan, from Google Earth Studio and manually annotate their paths to create ToSh-Data, which is used only for zero-shot evaluation. The construction procedure is shown in Fig.~\ref{fig:2}.

\subsubsection{Path annotation}
\begin{figure*}[!tp]
    \centering
    \includegraphics[width=\textwidth]{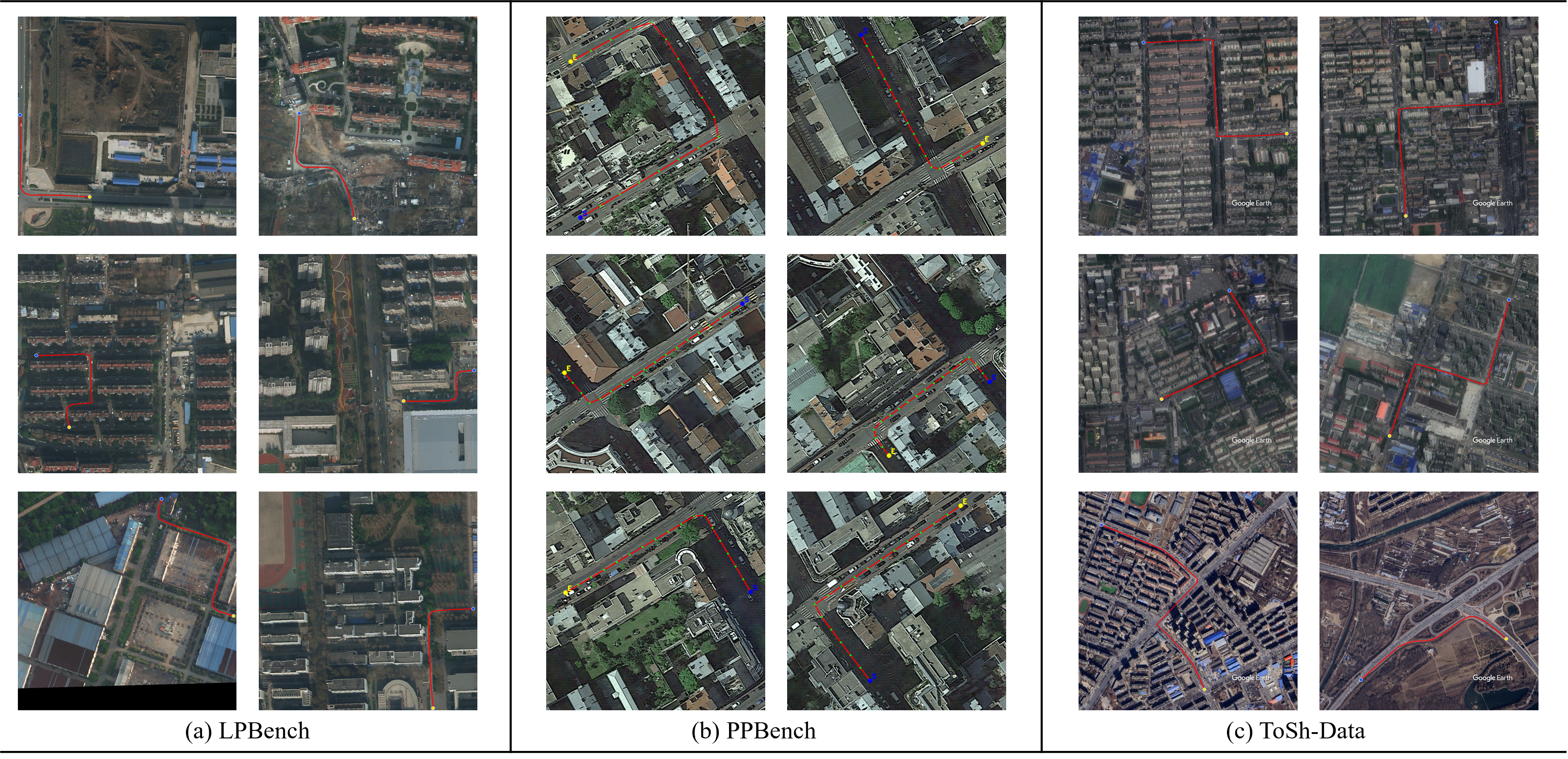}
    \caption{Ground-truth path annotations in (a) LPBench, (b) PPBench, and (c) ToSh-Data.}\label{fig:4}
\end{figure*}

For LoveDA, we combine manual annotation with generation by a Qwen model. Annotators first mark key path turning points rather than every point. Qwen then takes these points and the image to fill in intermediate points along the path. Finally, annotators correct the complete paths. Example ground-truth paths are shown in Fig.~\ref{fig:4}(a). This greatly reduces annotation time, for the first manual pass requires only about three to five key points per path on average.

For Paris, we first crop the original data into $1024\times1024$ images. As in pre-training data construction, we use scikit-image binary morphological skeletonization and sample the resulting road skeletons to obtain path coordinates, as shown in Fig.~\ref{fig:3}(b). Annotators then inspect the paths for all images and remove unreasonable samples, such as self-intersecting paths. The retained ground-truth paths are illustrated in Fig.~\ref{fig:4}(b).

\subsubsection{Constructing the three tasks}
For each reasonable path, we construct \CoarsePath{}, \FullPath{}, and \CoarseFull{} samples using the procedure for real pre-training data. The difference is that the more expensive and capable Qwen3.8-Max replaces Qwen3.7-Plus when generating \FullDesc{} text.

\subsubsection{Verification}
We add a verification step to ensure that \FullDesc{} and \Coarse{} descriptions agree. DeepSeek-v4-Flash~\cite{xu2026deepseek} receives \FullDesc{} text and extracts an ordered sequence of directional trends. It must select directions from the same eight template terms: left, right, upward, downward, upper left, lower left, upper right, and lower right. We compare the extracted sequence with the trends in \Coarse{} and identify samples containing directions in the extracted result that are absent from \Coarse{}. Annotators further inspect these inconsistencies. If \FullDesc{} contains a confirmed direction error, annotators interact with GPT-5.6 Sol~\cite{openai2026gpt56systemcard} to obtain a corrected \FullDesc{} description.
\begin{table*}[t]
\centering\caption{Numbers of path samples in each stratum of the three benchmarks under the two stratification strategies.}\label{tab:2}
\small\setlength{\tabcolsep}{3pt}
\begin{tabularx}{\textwidth}{llc*{8}{>{\centering\arraybackslash}X}}
\toprule
\multirow{2}{*}{Dataset} & \multirow{2}{*}{Split} & \multirow{2}{*}{\makecell{Path samples\\number}} &  \multicolumn{5}{c}{Trend count} & \multicolumn{3}{c}{Path pixel length} \\
 &  &  & 1 & 2 & 3 & 4 & 5 & $<600$ & 600--1000 & $>1000$ \\
\midrule
\multicolumn{11}{c}{\textbf{Benchmarks}} \\
\midrule
\multirow{2}{*}{\auditcell{2!A4}{LPBench}} & \auditcell{2!B4}{train} & \auditcell{2!C4}{1024} & \auditcell{2!D4}{54} & \auditcell{2!E4}{661} & \auditcell{2!F4}{246} & \auditcell{2!G4}{59} & \auditcell{2!H4}{4} & \auditcell{2!I4}{770} & \auditcell{2!J4}{239} & \auditcell{2!K4}{15} \\
 & \auditcell{2!B5}{test} & \auditcell{2!C5}{2047} & \auditcell{2!D5}{97} & \auditcell{2!E5}{1423} & \auditcell{2!F5}{423} & \auditcell{2!G5}{88} & \auditcell{2!H5}{16} & \auditcell{2!I5}{1509} & \auditcell{2!J5}{510} & \auditcell{2!K5}{28} \\
\multirow{2}{*}{\auditcell{2!A6}{PPBench}} & \auditcell{2!B6}{train} & \auditcell{2!C6}{744} & \auditcell{2!D6}{35} & \auditcell{2!E6}{525} & \auditcell{2!F6}{158} & \auditcell{2!G6}{26} & \auditcell{2!H6}{0} & \auditcell{2!I6}{0} & \auditcell{2!J6}{294} & \auditcell{2!K6}{450} \\
 & \auditcell{2!B7}{test} & \auditcell{2!C7}{1483} & \auditcell{2!D7}{73} & \auditcell{2!E7}{1066} & \auditcell{2!F7}{296} & \auditcell{2!G7}{48} & \auditcell{2!H7}{0} & \auditcell{2!I7}{0} & \auditcell{2!J7}{612} & \auditcell{2!K7}{871} \\
\midrule
\multicolumn{11}{c}{\textbf{Zero-shot data}} \\
\midrule
\auditcell{2!A9}{ToSh-Data} & \auditcell{2!B9}{test} & \auditcell{2!C9}{34} & \auditcell{2!D9}{0} & \auditcell{2!E9}{15} & \auditcell{2!F9}{18} & \auditcell{2!G9}{1} & \auditcell{2!H9}{0} & \auditcell{2!I9}{0} & \auditcell{2!J9}{6} & \auditcell{2!K9}{28} \\
\bottomrule\end{tabularx}
\end{table*}

This procedure produces LPBench from LoveDA and PPBench from Paris. LPBench contains 1708 images and 3071 paths, since an image may have multiple paths, and is divided into training and test splits. PPBench contains 2227 images and 2227 paths and is also divided into training and test splits.

For ToSh-Data, Google Earth Studio images of Tokyo are collected at a camera altitude of 1500 m and a tilt angle of $0^\circ$, yielding 15 images. Shenyang images are collected at 3000 m and $0^\circ$, yielding 19 images. We manually annotate paths and construct \Coarse{} using the real pre-training data procedure. For \FullDesc{}, annotators directly interact with GPT-5.6 Sol using the image and \Coarse{} text. All ToSh-Data samples evaluate zero-shot performance after pre-training. Ground-truth paths are shown in Fig.~\ref{fig:4}(c).

Table~\ref{tab:1} compares our three benchmarks with existing visual-language benchmarks. For LPBench and PPBench, we evaluate both zero-shot and fine-tuned performance. Zero-shot evaluation directly tests the pre-trained methods on the test splits. Fine-tuned evaluation first fine-tunes the pre-trained methods on the corresponding training split and then evaluates the test split. All three tasks in the LPBench and PPBench training splits participate in fine-tuning.

LPBench, PPBench, and ToSh-Data support stratified evaluation. Paths are divided into three pixel arc-length levels, $<600$, 600--1000, and $>1000$, and five directional trend-count levels, 1--5. Each level can be evaluated separately. We also evaluate \Coarse{} and \FullDesc{} inputs separately. Table~\ref{tab:2} reports path counts in every stratum.

\subsection{Metrics for Path Generation}

We design three metrics to quantify differences between a model's output path and the reference ground-truth path.

\subsubsection{Normalized Average Displacement Error (NADE)}
NADE measures the mean distance between corresponding ordered points on the predicted and ground-truth paths. Before computing it, predictions from all three architectures are rescaled to the $448\times448$ visual input size of PaliGemma 2 3B. Head-based and denoising-based outputs are directly rescaled from $[-1,1]$. NTP-based outputs are first mapped back to $[-1,1]$ and then to $448\times448$. Ground-truth paths normalized to $[-1,1]$ during training are rescaled in the same way. We then resample path points. We consider $\kappa$$=\{200,300,400,500,600,700,800\}$, a set of seven values controlling the resampling density, and compute metrics for each value. The selected value is multiplied by pixel arc length and divided by the diagonal length to determine the number of resampled points. Let $P$ and $G$ denote the predicted and ground-truth paths before resampling, and $\widetilde P$ and $\widetilde G$ denote their resampled versions. Equal arc-length sampling uses the resulting point count. For either original path $Q$, let $L(Q)$ be its pixel polyline length. Its resampled point count is given by \eqref{eq:1}.

\begin{equation}
    n_\kappa(Q)=\max\left\{20,\left\lceil\frac{\kappa L(Q)}{447\sqrt{2}}\right\rceil\right\}.\label{eq:1}
\end{equation}

For NADE, the ground-truth path determines the resampled point count $n_G$. Both paths are sampled at equal arc-length intervals using this count. Euclidean distances between corresponding points on $\widetilde P$ and $\widetilde G$ are averaged and normalized by the diagonal length $D$, as in \eqref{eq:2}.

\begin{equation}
    \mathrm{NADE}_\kappa(P,G)=\frac{1}{n_G D}\sum_{i=1}^{n_G}\left\|\widetilde P_i-\widetilde G_i\right\|_2.\label{eq:2}
\end{equation}

Diagonal normalization makes the metric independent of the visual encoder's input dimensions, allowing comparison between models with different visual input sizes.

\subsubsection{Normalized Chamfer Distance (NCD)}
NCD uses the same coordinate rescaling and resampling procedure as NADE. It instead calculates two distances: the mean nearest-point distance from the predicted path to ground truth and the mean nearest-point distance from ground truth to the predicted path. For the former, the predicted path length is substituted into \eqref{eq:1} to obtain $n_P$, while for the latter, the ground-truth length determines $n_G$. The two mean distances are averaged and normalized by the diagonal length. The pointwise Euclidean distance and NCD are defined in \eqref{eq:3} and \eqref{eq:4}, respectively.

\begin{equation}
    d_{ij}=\sqrt{(\widetilde P_{i,x}-\widetilde G_{j,x})^2+(\widetilde P_{i,y}-\widetilde G_{j,y})^2}.\label{eq:3}
\end{equation}

\begin{equation}
    \begin{aligned}\mathrm{NCD}_\kappa(P,G)=\frac{1}{2D}\Bigg[&\frac{1}{n_P}\sum_{i=1}^{n_P}\min_{1\le j\le n_G}d_{ij}\\+&\frac{1}{n_G}\sum_{j=1}^{n_G}\min_{1\le i\le n_P}d_{ij}\Bigg]\end{aligned}\label{eq:4}
\end{equation}

\subsubsection{Path Length Error (PLE)}
After rescaling predictions and ground-truth coordinates to $448\times448$, PLE further normalizes them to $[0,1]$, yielding $U_i$ and $V_i$ in \eqref{eq:5}. No resampling is required here. Predicted start and end points are replaced by their ground-truth values. We calculate the normalized predicted and ground-truth polyline lengths $L_U$ and $L_V$, as in \eqref{eq:6} and \eqref{eq:7}. Their absolute difference is divided by the larger length, giving PLE in \eqref{eq:8}, with $\varepsilon=10^{-8}$. Dividing by the larger length constrains the metric to $[0,1]$.

\begin{equation}
    U_i=\frac{P_i}{447},\quad V_i=\frac{G_i}{447},\quad i=1,\ldots,20.\label{eq:5}
\end{equation}

\begin{equation}
    L_U=\sum_{i=1}^{19}\sqrt{(U_{i+1,x}-U_{i,x})^2+(U_{i+1,y}-U_{i,y})^2},\label{eq:6}
\end{equation}

\begin{equation}
    L_V=\sum_{i=1}^{19}\sqrt{(V_{i+1,x}-V_{i,x})^2+(V_{i+1,y}-V_{i,y})^2}.\label{eq:7}
\end{equation}

\begin{equation}
    \mathrm{PLE}(P,G)=\frac{|L_U-L_V|}{\max\{L_U,L_V,\varepsilon\}}\label{eq:8}
\end{equation}

For all three metrics, predicted start and end points are first replaced with their ground-truth values before subsequent computation. Autoregressive models may produce outputs that violate the required format. We exclude three invalid cases: more coordinate points than required (20 by default), fewer points than required, and outputs containing non-numerical characters. Metrics and comparisons are calculated only over format-valid predictions, and format validity does not imply prediction correctness. Lower values are better for all three metrics.

\section{Experiments and Discussion}

We design several experiments to evaluate the three methods on \CoarsePath{} and \FullPath{}. Every experiment, including pre-training and supervised fine-tuning (SFT), uses four devices, with a batch size of one per device and gradient accumulation over two steps. All parameters of each architecture are fine-tuned. All training uses eager attention operations to retain the modified attention computation of PaliGemma 2 3B. All reported results use $\kappa$$=800$, since metric values gradually stabilize as $\kappa$ increases. All metric values reported in the following experiments are expressed as percentages.

\subsection{Pre-training Settings}
The three path-related tasks and VRSBench VQA samples are loaded sequentially at a ratio of 3:1. One complete traversal of VRSBench therefore loads 343,252 samples and requires 42,907 update steps under the above configuration. Head-based and Denoising-based methods use three traversals, i.e., 128,721 steps. The NTP-based method uses four traversals, i.e., 171,628 steps, to promote convergence.

\subsection{Supervised Fine-tuning Settings}

After task-specific pre-training, only GK-v2 is used to introduce general knowledge. On LPBench, benchmark training samples and GK-v2 samples are loaded at a 3:1 ratio. One traversal of GK-v2 requires 3000 update steps. The w/o GK experiment also uses 3000 steps to study the effect of the data mixture with the update count held constant. On PPBench, the loading ratio is 2:1, requiring 2250 steps for one GK-v2 traversal, where the w/o GK setting also uses 2250 steps.

For SFT without RemotePath-175k task-specific pre-training, we separately introduce GK-v1 and GK-v2. These experiments retain the original PaliGemma 2 3B pre-trained weights. With GK-v2 (w/ GK-v2), loading ratios remain 3:1 for LPBench and 2:1 for PPBench, requiring 3000 and 2250 update steps, respectively. With GK-v1 (w/ GK-v1), we retain the same number of benchmark training samples per complete GK traversal as with GK-v2, while visiting all GK-v1 samples once. This allows us to examine whether introducing more general knowledge improves performance. Loading ratios for benchmark and GK-v1 are then 3:4 for LPBench and 1:2 for PPBench, requiring 5250 and 4500 update steps, respectively. The w/o GK setting retains the GK-v2 step counts: 3000 on LPBench and 2250 on PPBench. Here w/o GK means that no additional GK dataset is mixed in.

\subsection{Comparison of the Three Architectures and Advanced Commercial Models}
We evaluate zero-shot performance after pre-training on LPBench, PPBench, and ToSh-Data, and performance after fine-tuning on LPBench and PPBench, as reported in Table~\ref{tab:3}. We also evaluate seven advanced commercial models in the zero-shot setting on all three benchmarks: Qwen3.7-Plus~\cite{qwen37plus}, Qwen3.8-Flash~\cite{qwen2026design,qwen3.8flashnext}, GLM-5.3-Flash~\cite{glm5team2026glm5vibecodingagentic,zai2026glm53flash}, Gemini 3.6 Flash~\cite{google2026gemini36flash}, GPT-5.6 Sol~\cite{openai2026gpt56systemcard}, GPT-5.6 Luna~\cite{openai2026gpt56systemcard}, and Claude Sonnet 5~\cite{anthropic2026sonnet5}. Table~\ref{tab:4} reports the counts and proportions of format-valid NTP-based predictions in the three settings.

\begin{table*}[t]
\centering\caption{Overall performance after RemotePath-175k pre-training: zero-shot evaluation and fine-tuning w/o GK or w/ GK-v2, compared with seven recent commercial VLMs. All metrics are percentages, and lower is better.}\label{tab:3}
\small\setlength{\tabcolsep}{3pt}
\begin{tabular*}{\textwidth}{@{\extracolsep{\fill}}p{3cm}p{1.8cm}rrrrrrrrr@{}}
\toprule
\multicolumn{11}{c}{\textbf{\Coarse{} input}} \\
\midrule
\multirow{2}{*}{Model} & \multirow{2}{*}{Test mode} & \multicolumn{3}{c}{LPBench} & \multicolumn{3}{c}{PPBench} & \multicolumn{3}{c}{ToSh-Data} \\
 & & NADE & NCD & PLE & NADE & NCD & PLE & NADE & NCD & PLE \\
\midrule
Qwen3.7-Plus & zero-shot & \auditcell{3!C4}{2.62} & \auditcell{3!D4}{1.88} & \auditcell{3!E4}{9.97} & \auditcell{3!I4}{6.33} & \auditcell{3!J4}{4.57} & \auditcell{3!K4}{12.60} & \auditcell{3!O4}{6.15} & \auditcell{3!P4}{4.58} & \auditcell{3!Q4}{13.27}  \\
Qwen3.8-Flash & zero-shot & \auditcell{3!C5}{2.29} & \auditcell{3!D5}{1.74} & \auditcell{3!E5}{9.32} & \auditcell{3!I5}{5.27} & \auditcell{3!J5}{3.94} & \auditcell{3!K5}{11.21} & \auditcell{3!O5}{5.82} & \auditcell{3!P5}{4.27} & \auditcell{3!Q5}{11.65}  \\
GLM-5.3-Flash & zero-shot & \auditcell{3!C6}{1.35} & \auditcell{3!D6}{1.09} & \auditcell{3!E6}{5.53} & \auditcell{3!I6}{3.40} & \auditcell{3!J6}{2.56} & \auditcell{3!K6}{7.31} & \auditcell{3!O6}{3.92} & \auditcell{3!P6}{2.90} & \auditcell{3!Q6}{6.39}  \\
Gemini 3.6 Flash & zero-shot & \auditcell{3!C7}{1.10} & \auditcell{3!D7}{0.83} & \auditcell{3!E7}{4.74} & \auditcell{3!I7}{2.85} & \auditcell{3!J7}{2.00} & \auditcell{3!K7}{6.36} & \auditcell{3!O7}{2.02} & \auditcell{3!P7}{1.44} & \auditcell{3!Q7}{4.31}  \\
GPT-5.6 Sol & zero-shot & \auditcell{3!C8}{1.16} & \auditcell{3!D8}{0.94} & \auditcell{3!E8}{5.08} & \auditcell{3!I8}{3.02} & \auditcell{3!J8}{2.28} & \auditcell{3!K8}{6.70} & \auditcell{3!O8}{2.69} & \auditcell{3!P8}{1.99} & \auditcell{3!Q8}{6.07}  \\
GPT-5.6 Luna & zero-shot & \auditcell{3!C9}{1.21} & \auditcell{3!D9}{0.97} & \auditcell{3!E9}{5.17} & \auditcell{3!I9}{3.00} & \auditcell{3!J9}{2.28} & \auditcell{3!K9}{6.57} & \auditcell{3!O9}{3.02} & \auditcell{3!P9}{2.28} & \auditcell{3!Q9}{7.10}  \\
Claude Sonnet 5 & zero-shot & \auditcell{3!C10}{1.84} & \auditcell{3!D10}{1.47} & \auditcell{3!E10}{8.49} & \auditcell{3!I10}{4.02} & \auditcell{3!J10}{3.08} & \auditcell{3!K10}{9.07} & \auditcell{3!O10}{3.50} & \auditcell{3!P10}{2.61} & \auditcell{3!Q10}{6.92}  \\
\midrule
\multirow{3}{*}{NTP-based (ours)} & zero-shot & \auditcell{3!C11}{6.33} & \auditcell{3!D11}{3.09} & \auditcell{3!E11}{34.37} & \auditcell{3!I11}{2.61} & \auditcell{3!J11}{1.97} & \auditcell{3!K11}{4.09} & \auditcell{3!O11}{3.55} & \auditcell{3!P11}{2.44} & \auditcell{3!Q11}{5.11}  \\
 & w/o GK & \auditcell{3!C12}{2.42} & \auditcell{3!D12}{1.70} & \auditcell{3!E12}{10.55} & \auditcell{3!I12}{4.84} & \auditcell{3!J12}{3.43} & \auditcell{3!K12}{9.83} & \auditcell{3!O12}{---} & \auditcell{3!P12}{---} & \auditcell{3!Q12}{---}  \\
 & w/ GK-v2 & \auditcell{3!C13}{2.10} & \auditcell{3!D13}{1.53} & \auditcell{3!E13}{9.46} & \auditcell{3!I13}{3.47} & \auditcell{3!J13}{2.57} & \auditcell{3!K13}{6.55} & \auditcell{3!O13}{---} & \auditcell{3!P13}{---} & \auditcell{3!Q13}{---}  \\
\midrule
\multirow{3}{*}{Head-based (ours)} & zero-shot & \auditcell{3!C14}{2.24} & \auditcell{3!D14}{1.64} & \auditcell{3!E14}{11.78} & \auditcell{3!I14}{2.26} & \auditcell{3!J14}{1.76} & \auditcell{3!K14}{4.28} & \auditcell{3!O14}{2.92} & \auditcell{3!P14}{2.13} & \auditcell{3!Q14}{6.64}  \\
 & w/o GK & \auditcell{3!C15}{1.40} & \auditcell{3!D15}{1.10} & \auditcell{3!E15}{5.99} & \auditcell{3!I15}{2.45} & \auditcell{3!J15}{1.83} & \auditcell{3!K15}{5.27} & \auditcell{3!O15}{---} & \auditcell{3!P15}{---} & \auditcell{3!Q15}{---}  \\
 & w/ GK-v2 & \auditcell{3!C16}{1.40} & \auditcell{3!D16}{1.11} & \auditcell{3!E16}{5.85} & \auditcell{3!I16}{2.23} & \auditcell{3!J16}{1.67} & \auditcell{3!K16}{4.63} & \auditcell{3!O16}{---} & \auditcell{3!P16}{---} & \auditcell{3!Q16}{---}  \\
\midrule
\multirow{3}{*}{Denoising-based (ours)} & zero-shot & \auditcell{3!C17}{2.63} & \auditcell{3!D17}{1.67} & \auditcell{3!E17}{25.70} & \auditcell{3!I17}{2.30} & \auditcell{3!J17}{1.72} & \auditcell{3!K17}{6.33} & \auditcell{3!O17}{2.56} & \auditcell{3!P17}{1.81} & \auditcell{3!Q17}{5.32}  \\
 & w/o GK & \auditcell{3!C18}{1.68} & \auditcell{3!D18}{1.11} & \auditcell{3!E18}{19.82} & \auditcell{3!I18}{2.18} & \auditcell{3!J18}{1.57} & \auditcell{3!K18}{5.35} & \auditcell{3!O18}{---} & \auditcell{3!P18}{---} & \auditcell{3!Q18}{---}  \\
 & w/ GK-v2 & \auditcell{3!C19}{1.72} & \auditcell{3!D19}{1.15} & \auditcell{3!E19}{19.79} & \auditcell{3!I19}{2.11} & \auditcell{3!J19}{1.55} & \auditcell{3!K19}{5.01} & \auditcell{3!O19}{---} & \auditcell{3!P19}{---} & \auditcell{3!Q19}{---}  \\
\midrule
\multicolumn{11}{c}{\textbf{\FullDesc{} input}} \\
\midrule
\multirow{2}{*}{Model} & \multirow{2}{*}{Test mode} & \multicolumn{3}{c}{LPBench} & \multicolumn{3}{c}{PPBench} & \multicolumn{3}{c}{ToSh-Data} \\
 & & NADE & NCD & PLE & NADE & NCD & PLE & NADE & NCD & PLE \\
\midrule
Qwen3.7-Plus & zero-shot & \auditcell{3!F4}{3.64} & \auditcell{3!G4}{2.50} & \auditcell{3!H4}{14.60} & \auditcell{3!L4}{6.63} & \auditcell{3!M4}{4.76} & \auditcell{3!N4}{13.34} & \auditcell{3!R4}{7.48} & \auditcell{3!S4}{5.57} & \auditcell{3!T4}{13.45}  \\
Qwen3.8-Flash & zero-shot & \auditcell{3!F5}{3.23} & \auditcell{3!G5}{2.36} & \auditcell{3!H5}{13.68} & \auditcell{3!L5}{5.81} & \auditcell{3!M5}{4.28} & \auditcell{3!N5}{12.93} & \auditcell{3!R5}{8.55} & \auditcell{3!S5}{6.19} & \auditcell{3!T5}{16.41}  \\
GLM-5.3-Flash & zero-shot & \auditcell{3!F6}{2.14} & \auditcell{3!G6}{1.62} & \auditcell{3!H6}{9.95} & \auditcell{3!L6}{3.94} & \auditcell{3!M6}{2.96} & \auditcell{3!N6}{8.86} & \auditcell{3!R6}{4.90} & \auditcell{3!S6}{3.63} & \auditcell{3!T6}{11.71}  \\
Gemini 3.6 Flash & zero-shot & \auditcell{3!F7}{1.98} & \auditcell{3!G7}{1.29} & \auditcell{3!H7}{8.77} & \auditcell{3!L7}{3.34} & \auditcell{3!M7}{2.27} & \auditcell{3!N7}{7.64} & \auditcell{3!R7}{3.20} & \auditcell{3!S7}{2.20} & \auditcell{3!T7}{5.58}  \\
GPT-5.6 Sol & zero-shot & \auditcell{3!F8}{1.90} & \auditcell{3!G8}{1.50} & \auditcell{3!H8}{9.41} & \auditcell{3!L8}{4.00} & \auditcell{3!M8}{2.96} & \auditcell{3!N8}{9.98} & \auditcell{3!R8}{4.88} & \auditcell{3!S8}{3.63} & \auditcell{3!T8}{14.32}  \\
GPT-5.6 Luna & zero-shot & \auditcell{3!F9}{2.31} & \auditcell{3!G9}{1.82} & \auditcell{3!H9}{10.29} & \auditcell{3!L9}{4.34} & \auditcell{3!M9}{3.27} & \auditcell{3!N9}{10.37} & \auditcell{3!R9}{6.08} & \auditcell{3!S9}{4.58} & \auditcell{3!T9}{15.33}  \\
Claude Sonnet 5 & zero-shot & \auditcell{3!F10}{3.32} & \auditcell{3!G10}{2.62} & \auditcell{3!H10}{15.88} & \auditcell{3!L10}{6.51} & \auditcell{3!M10}{4.99} & \auditcell{3!N10}{16.31} & \auditcell{3!R10}{7.13} & \auditcell{3!S10}{5.32} & \auditcell{3!T10}{19.61}  \\
\midrule
\multirow{3}{*}{NTP-based (ours)} & zero-shot & \auditcell{3!F11}{8.33} & \auditcell{3!G11}{4.27} & \auditcell{3!H11}{40.88} & \auditcell{3!L11}{4.66} & \auditcell{3!M11}{3.45} & \auditcell{3!N11}{9.06} & \auditcell{3!R11}{5.87} & \auditcell{3!S11}{4.21} & \auditcell{3!T11}{8.62}  \\
 & w/o GK & \auditcell{3!F12}{2.83} & \auditcell{3!G12}{1.95} & \auditcell{3!H12}{12.69} & \auditcell{3!L12}{5.93} & \auditcell{3!M12}{4.08} & \auditcell{3!N12}{12.09} & \auditcell{3!R12}{---} & \auditcell{3!S12}{---} & \auditcell{3!T12}{---}  \\
 & w/ GK-v2 & \auditcell{3!F13}{2.58} & \auditcell{3!G13}{1.82} & \auditcell{3!H13}{11.96} & \auditcell{3!L13}{4.86} & \auditcell{3!M13}{3.46} & \auditcell{3!N13}{9.95} & \auditcell{3!R13}{---} & \auditcell{3!S13}{---} & \auditcell{3!T13}{---}  \\
\midrule
\multirow{3}{*}{Head-based (ours)} & zero-shot & \auditcell{3!F14}{4.08} & \auditcell{3!G14}{2.74} & \auditcell{3!H14}{23.22} & \auditcell{3!L14}{3.32} & \auditcell{3!M14}{2.52} & \auditcell{3!N14}{7.06} & \auditcell{3!R14}{4.81} & \auditcell{3!S14}{3.71} & \auditcell{3!T14}{11.02}  \\
 & w/o GK & \auditcell{3!F15}{1.66} & \auditcell{3!G15}{1.30} & \auditcell{3!H15}{7.85} & \auditcell{3!L15}{2.99} & \auditcell{3!M15}{2.20} & \auditcell{3!N15}{6.86} & \auditcell{3!R15}{---} & \auditcell{3!S15}{---} & \auditcell{3!T15}{---}  \\
 & w/ GK-v2 & \auditcell{3!F16}{1.71} & \auditcell{3!G16}{1.33} & \auditcell{3!H16}{8.20} & \auditcell{3!L16}{2.94} & \auditcell{3!M16}{2.16} & \auditcell{3!N16}{6.77} & \auditcell{3!R16}{---} & \auditcell{3!S16}{---} & \auditcell{3!T16}{---}  \\
\midrule
\multirow{3}{*}{Denoising-based (ours)} & zero-shot & \auditcell{3!F17}{4.43} & \auditcell{3!G17}{2.75} & \auditcell{3!H17}{31.98} & \auditcell{3!L17}{3.13} & \auditcell{3!M17}{2.29} & \auditcell{3!N17}{7.01} & \auditcell{3!R17}{4.24} & \auditcell{3!S17}{3.12} & \auditcell{3!T17}{7.92}  \\
 & w/o GK & \auditcell{3!F18}{1.92} & \auditcell{3!G18}{1.29} & \auditcell{3!H18}{19.51} & \auditcell{3!L18}{2.54} & \auditcell{3!M18}{1.80} & \auditcell{3!N18}{5.96} & \auditcell{3!R18}{---} & \auditcell{3!S18}{---} & \auditcell{3!T18}{---}  \\
 & w/ GK-v2 & \auditcell{3!F19}{2.03} & \auditcell{3!G19}{1.38} & \auditcell{3!H19}{19.37} & \auditcell{3!L19}{2.64} & \auditcell{3!M19}{1.88} & \auditcell{3!N19}{5.98} & \auditcell{3!R19}{---} & \auditcell{3!S19}{---} & \auditcell{3!T19}{---}  \\
\bottomrule
\end{tabular*}
\end{table*}

\begin{table*}[t]
\centering\caption{Numbers of format-valid predictions of the NTP-based method after pre-training, under zero-shot evaluation and fine-tuning w/o GK or w/ GK-v2 on LPBench nand PPBench.}\label{tab:4}
\small\setlength{\tabcolsep}{3pt}
\begin{tabular}{llcccccc}
\toprule
\multirow{2}{*}{Model} & \multirow{2}{*}{Test mode} &  \multicolumn{2}{c}{LPBench} & \multicolumn{2}{c}{PPBench} & \multicolumn{2}{c}{ToSh-Data} \\
 &  & \Coarse{} & \FullDesc{} & \Coarse{} & \FullDesc{} & \Coarse{} & \FullDesc{} \\
\midrule
\multirow{3}{*}{NTP-based} & zero-shot & \auditcell{4!D3}{1910/2047(93.3\%)} & \auditcell{4!E3}{1901/2047(92.9\%)} & \auditcell{4!F3}{1451/1483(97.8\%)} & \auditcell{4!G3}{1439/1483(97.0\%)} & \auditcell{4!H3}{33/34(97.1\%)} & \auditcell{4!I3}{34/34(100\%)} \\
 & w/o GK & \auditcell{4!D4}{1613/2047(78.8\%)} & \auditcell{4!E4}{1563/2047(76.4\%)} & \auditcell{4!F4}{1206/1483(81.3\%)} & \auditcell{4!G4}{1205/1483(81.3\%)} & \auditcell{4!H4}{---} & \auditcell{4!I4}{---} \\
 & w/ GK-v2 & \auditcell{4!D5}{1825/2047(89.2\%)} & \auditcell{4!E5}{1787/2047(87.3\%)} & \auditcell{4!F5}{1317/1483(88.8\%)} & \auditcell{4!G5}{1305/1483(88.0\%)} & \auditcell{4!H5}{---} & \auditcell{4!I5}{---} \\
\bottomrule\end{tabular}
\end{table*}

\subsubsection{Observation 1: fine-tuning versus zero-shot evaluation}
On LPBench, fine-tuning all three methods, with or without GK, reduces every metric under both \Coarse{} and \FullDesc{} compared with zero-shot evaluation. On PPBench, the pattern varies. For NTP-based generation, every metric under both inputs is higher after fine-tuning. For the head-based method, \Coarse{} metrics show both increases and decreases, whereas all \FullDesc{} metrics decrease. For the denoising-based method, all metrics decrease under both inputs.

One explanation for the consistent LPBench improvement and less consistent PPBench improvement is the difference in semantic complexity. PPBench mainly contains wide roads and regularly shaped buildings, with relatively homogeneous samples. Fine-tuning on PPBench may restrict the general knowledge of the pre-trained model, impair its understanding of complex scenes, and reduce generalization. Paths in LPBench are not limited to regular roads: some pass through traversable spaces between buildings or along building edges. Fine-tuning can teach path characteristics absent from pre-training and improve performance in these complex scenes.

The architecture-specific PPBench patterns may have the following explanations. First, NTP-based inference uses student forcing and may accumulate errors, resulting in higher metrics under both inputs after fine-tuning. Second, the head-based model does not accumulate errors in this way, so its degradation is less severe: there is some degradation under \Coarse{} and only a small improvement under \FullDesc{}. Third, in denoising-based method, the DiT expert receives ground-truth paths mixed with gaussian noise at randomly sampled times and predicts a velocity. Its inputs are therefore diverse, rather than simply the original homogeneous PPBench samples. This offsets the damage to pre-trained complex-scene understanding that may arise from directly learning homogeneous data. At the same time, supervision on PPBench teaches new domain patterns. The denoising-based method consequently outperforms its zero-shot setting after fine-tuning, with or without GK.

\subsubsection{Observation 2: introducing GK-v2 during fine-tuning}
On both LPBench and PPBench, all NTP-based metrics are lower w/ GK-v2 than w/o GK, although both settings remain worse than zero-shot evaluation on PPBench. For the head-based method, all metrics improve with GK only on PPBench, while this pattern does not hold on LPBench. For the denoising-based method, results are mixed on LPBench and under \FullDesc{} on PPBench. Only under \Coarse{} on PPBench does GK-v2 reduce every metric.

For the NTP-based method, GK reduces overfitting to benchmark training splits and injects broader knowledge. It improves all LPBench and PPBench metrics relative to  fine-tuning only with traing split in benchmark, although PPBench performance remains below the zero-shot level. This suggests that the method tends to require more training data: broader knowledge can improve next-token predictions and directly reduce later error accumulation. GK-v2 also increases the number of format-valid test predictions, suggesting that its path generation samples reinforce the output format, though validity remains below zero-shot levels. For \Coarse{}/\FullDesc{}, LPBench valid proportions are 93.3\%/92.9\% in the zero-shot setting, fall to 78.8\%/76.4\% w/o GK, and recover to 89.2\%/87.3\% w/ GK-v2. The corresponding PPBench values are 97.8\%/97.0\%, 81.3\%/81.3\%, and 88.8\%/88.0\%, as shown in Table~\ref{tab:4}.

For head-based and denoising-based methods, adding GK on LPBench does not give consistent improvements over training only on its training split, unlike the NTP-based result. Fine-tuning on only bechmark training split already performs well, and GK generally leaves performance near the previous level. LPBench training data already provide valuable, complex, multimodal semantics, while the additional general data contribute little further complex semantic information. On PPBench under \FullDesc{}, GK-v2 increases all three denoising-based metrics by at most 0.1, whereas all three head-based metrics improve by at most 0.09. Thus, neither architecture achieves a large improvement under \FullDesc{}. This may reflect PPBench's wide roads, regular buildings, and simple semantics, which the models already understand. The generally poorer performance under \FullDesc{} than under \Coarse{} may result from redundant information in \FullDesc{}, and GK-v2 descriptions contain the same redundancy, so adding GK therefore does not improve PPBench \FullDesc{} performance substantially. Under \Coarse{}, both methods improve relatively consistently with GK, which may enhance perception of direction and scale.

\subsubsection{Comparison with commercial models}
On LPBench, zero-shot metric values of our architectures are generally higher than those of most advanced commercial baselines, with lower values on only a few metrics. After fine-tuning, however, every head-based metrics under both \Coarse{} and \FullDesc{} are lower than those of Qwen3.7-Plus, Qwen3.8-Flash, and Claude Sonnet 5. On PPBench, head-based and denoising-based methods perform well in the zero-shot setting. The head-based method has lower values on every metric under both inputs than every commercial model except Gemini 3.6 Flash. Relative to Gemini 3.6 Flash, only its \FullDesc{} NCD is higher, by 0.25, while all other metrics are still lower. The denoising-based method follows the same pattern: only \FullDesc{} NCD is higher than Gemini 3.6 Flash, by 0.02, and all other metrics are lower than those of all commercial baselines. The NTP-based method has lower zero-shot metrics than all commercial models under \Coarse{} and lower values than some commercial baselines on some \FullDesc{} metrics. On ToSh-Data, the denoising-based method generally has the lowest values among our methods and outperforms every commercial model except Gemini 3.6 Flash on all metrics. NTP-based values are generally the highest among our methods, but remain lower than Qwen3.7-Plus and Qwen3.8-Flash on every metric.

These comparisons suggest that the constructed pre-training dataset has a degree of generality: training on it enables our three architectures to outperform advanced commercial models in some settings. The three architectures trained on the same dataset differ in performance, reflecting their architectural differences. Overall, head-based and denoising-based methods provide better zero-shot performance than NTP-based generation. Given the success of NTP in other domains, we believe that increasing pre-training data scale and diversity may further unlock its potential.

\begin{figure*}[!t]
    \centering
    \includegraphics[width=\textwidth]{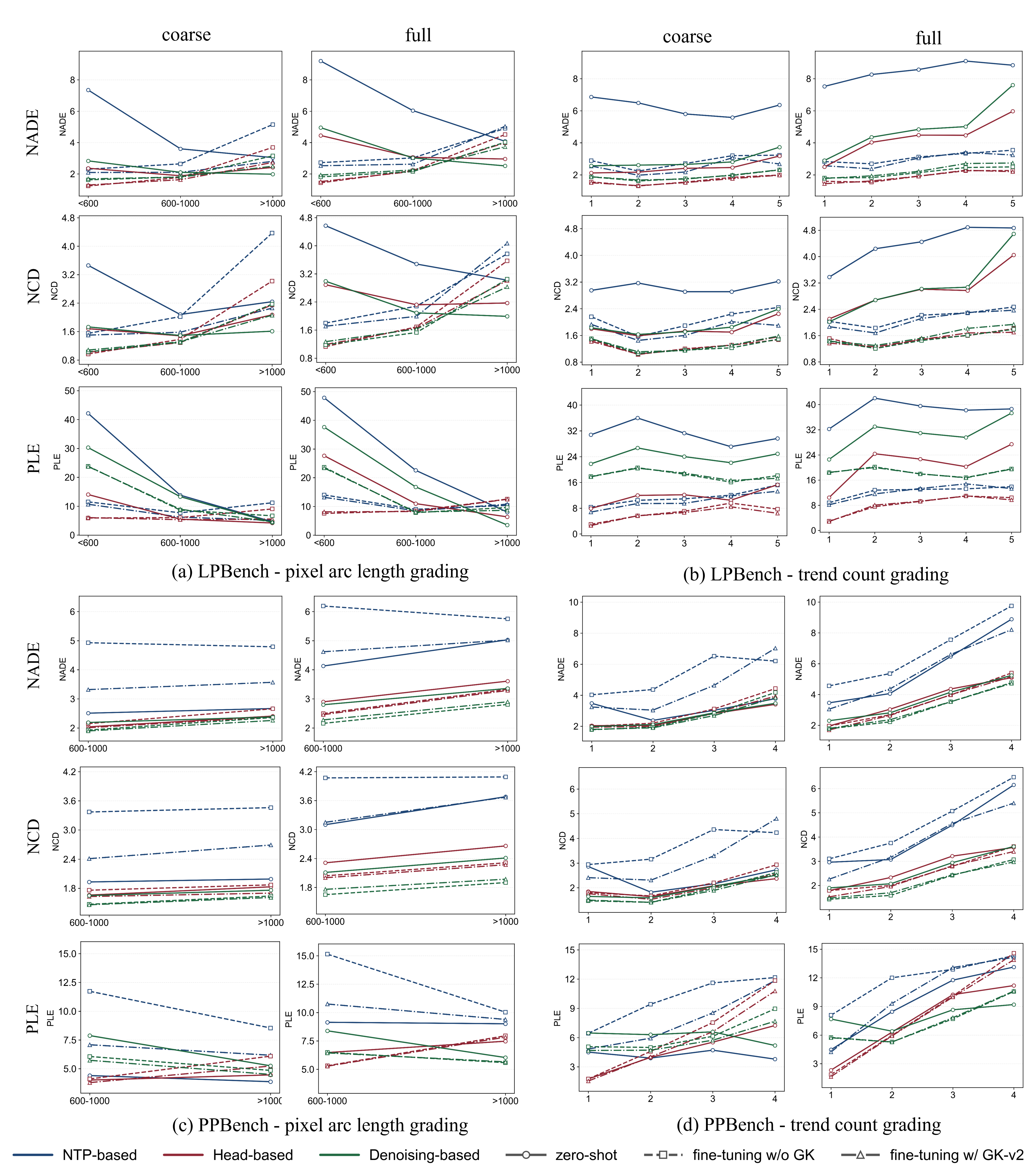}
    \caption{Stratified evaluation of all metrics under \Coarse{} and \FullDesc{} on LPBench and PPBench for the three methods: zero-shot evaluation, fine-tuning w/o GK, and fine-tuning w/ GK-v2.}\label{fig:5}
\end{figure*}

\subsection{Stratified Evaluation on LPBench and PPBench}
The benchmarks support stratification by pixel arc length and directional trend count. We apply both strategies under \Coarse{} and \FullDesc{}. Stratification reveals performance differences between sample groups and permits a more detailed study. We compare only our three architectures under zero-shot evaluation and fine-tuning w/ GK-v2 or w/o GK, as shown in Fig.~\ref{fig:5}.

\subsubsection{Observation 1: pixel arc-length strata on LPBench}
For zero-shot NCD under \FullDesc{}, all three methods have higher values in the $<600$ stratum than in 600--1000 and $>1000$. Under \Coarse{}, the $<600$ stratum has the highest zero-shot NCD for NTP-based and denoising-based methods and the second-highest for the head-based method. After fine-tuning w/ GK-v2 or w/o GK, all three methods have lower NCD in $<600$ than in the other two strata under both inputs.
For zero-shot PLE, all three methods have higher values in $<600$ than in the other two strata under both inputs. After fine-tuning w/ GK-v2 or w/o GK, NTP-based and denoising-based PLE remain higher in $<600$ under both inputs. For the head-based method, this occurs only under \Coarse{} w/ GK-v2. Under \Coarse{} w/o GK and under \FullDesc{} with either setting, its PLE is lower in $<600$ than in 600--1000 and $>1000$.

The zero-shot NCD pattern may mainly reflect the relatively small number of pre-training paths shorter than 600 pixels. The methods learn longer-path generation more sufficiently than short-path generation. After LPBench fine-tuning, NCD is lowest in $<600$ under both inputs, with or without GK-v2. Short paths account for about 75.2\% of the LPBench training split, allowing the models to learn their generation more sufficiently.

The larger PLE values on short paths are related to the metric definition. PLE divides the absolute pixel arc-length difference by the greater of the predicted and true lengths. For the same scale of point jitter, detour, or shortcut, the error occupies a greater fraction of a short path than a long one. NCD measures mean nearest-point distances between paths and is therefore less directly related to path length than PLE.

In zero-shot evaluation, the shortage of $<600$ pre-training paths and the greater sensitivity of PLE to the same error on short paths together produce larger PLE in $<600$ under both inputs. After SFT, with or without GK-v2, the relative PLE values across length strata depend both on this metric property and on the architecture.

\begin{figure*}[!t]
    \centering
    \includegraphics[width=\textwidth]{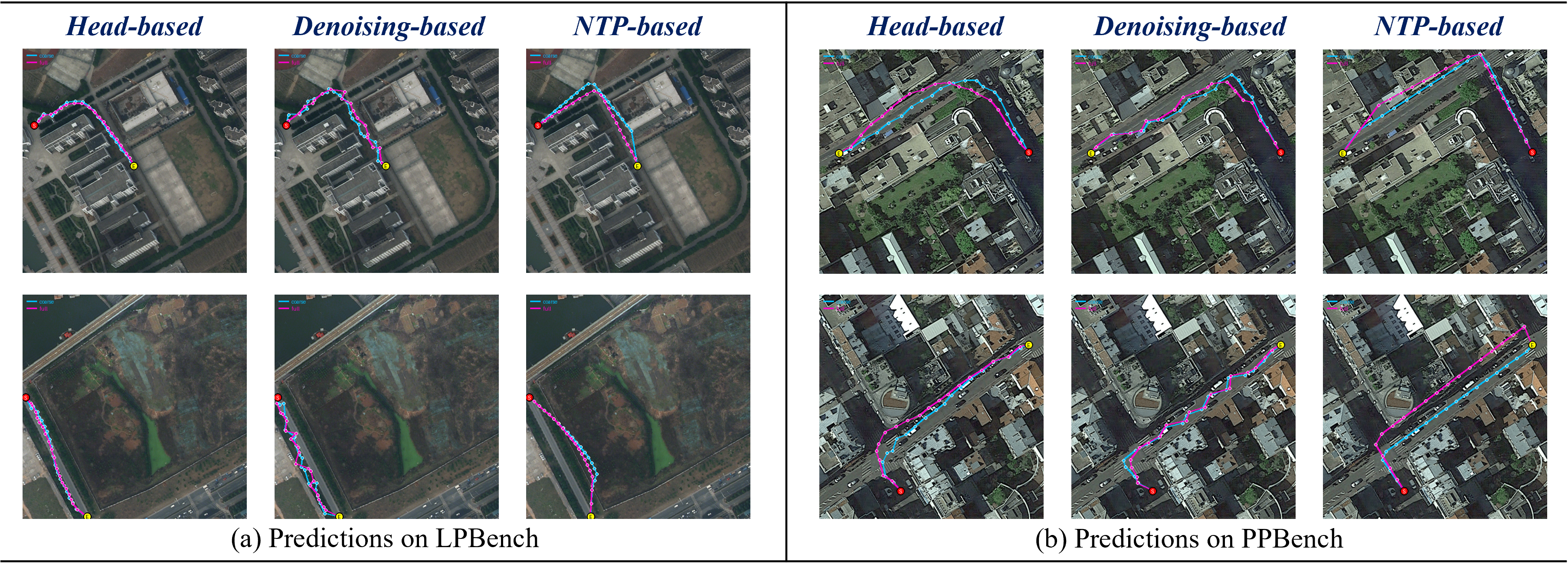}
    \caption{Predicted paths from the three methods on (a) LPBench and (b) PPBench. Architectural differences result in different path-shape characteristics.}\label{fig:6}
\end{figure*}

For NTP-based and denoising-based methods, PLE in $<600$ remains higher than in 600--1000 and $>1000$ under both inputs. Fine-tuned NTP-based paths generally do not show substantial point jitter. However, error accumulation in next-token prediction may produce detours or incorrect overall paths, as shown in Fig.~\ref{fig:6}. PLE amplifies such errors for short paths. Fine-tuned denoising-based paths instead show strong point jitter around ground truth. The method learns the velocity field for all states along the trajectory from noise to the final path. It learns the difference between ground truth and noise rather than directly learning the final shape and relationships between path points. Its final predictions therefore lack a direct understanding of point relationships and contain substantial jitter, as illustrated in Fig.~\ref{fig:6}.

For the head-based method, PLE in $<600$ exceeds the other strata only under \Coarse{} w/ GK-v2, by at most 0.7. In all other cases it is lower. Together with the lower fine-tuned NCD in $<600$ under both inputs, this can be understood from the visualizations. The head-based method has neither the frequent detours of NTP-based generation nor the strong point jitter of denoising. It does not accumulate inference errors as NTP does and directly learns ground-truth coordinates, giving a direct understanding of path shape and point relationships. Thus, both NCD and PLE are relatively small for short paths under both inputs.

\subsubsection{Observation 2: trend-count strata on LPBench}
For NCD under \Coarse{}, trend count 5 has the highest values in most cases, and count 2 has the lowest in most cases. Under \FullDesc{}, count 5 also has the highest NCD in most cases. All methods have their lowest zero-shot NCD at count 1 under \FullDesc{}. After SFT, all have their lowest NCD at count 2, with count 1 usually second-lowest. For PLE under both inputs, count 1 gives the lowest values in most zero-shot and SFT cases.

The generally highest NCD at count 5 may reflect the limited number of such samples in pre-training and the extremely small number in LPBench. Training for these paths is therefore insufficient. More trends also require predicting turning locations and several extension directions, making the prediction pattern more complex. Under \FullDesc{}, zero-shot NCD is lowest at count 1 because no direction change must be predicted: only length is needed, making this the simplest pattern. After fine-tuning under \FullDesc{} and in most \Coarse{} cases, count 2 gives the lowest NCD because it is the most common count in the LPBench training split and is learned most sufficiently.

For PLE, more trends can introduce smooth predictions around turns and inaccurate turning locations, producing detours or shortcuts and increasing length error. Count 1 avoids these turning-related errors and therefore has the lowest PLE in most cases.

\subsubsection{Observation 3: pixel arc-length strata on PPBench}
Under both inputs, all three methods have lower NCD in 600--1000 than in $>1000$, for zero-shot evaluation and fine-tuning with or without GK-v2. For PLE, NTP-based and denoising-based methods always have higher values in 600--1000 than in $>1000$ under both inputs, while head-based method has lower PLE in 600--1000 than in $>1000$ under both inputs.

Both length strata are well represented in pre-training and the PPBench training split, so performance differences due to sample counts can be neglected. Longer paths require more complex prediction patterns and greater spatial understanding, giving higher NCD.

As on LPBench, the NTP-based and denoising-based PLE pattern mainly reflects the metric's sensitivity to a disturbance of the same magnitude on a shorter path. The head-based method directly learns path coordinates through its decoder and successfully learns global directions. However, it may not accurately predict points near important locations such as turns or the precise direction of each segment. Smooth turning predictions can create shortcuts or detours, as shown in Fig.~\ref{fig:6}. On longer paths, the resulting length changes are more pronounced, so PLE increases with path length.

\subsubsection{Observation 4: trend-count strata on PPBench}
For NCD under \Coarse{}, count 2 gives the lowest values in most cases and count 4 gives the highest in most cases. Under \FullDesc{}, all three methods have the lowest NCD at count 1 and the highest at count 4 in every setting. For PLE under \Coarse{}, count 1 is lowest and count 4 highest in most cases. Under \FullDesc{}, count 1 is lowest in most cases, while count 4 is highest in every setting.

The PPBench training split contains far more count 2 than count 4 samples. Training imbalance may therefore explain the generally highest NCD at count 4 and the relatively low \Coarse{} NCD at count 2. Under \FullDesc{}, count 1 has the lowest NCD because it requires no turn or trend change and has a simpler prediction pattern.

The PLE explanation is similar to that for LPBench. More directional trends involve more turns, where all three methods are prone to errors such as smooth transitions or misplaced turning points. These produce shortcuts or detours. As turns increase, such errors become more frequent and increase the path-length difference relative to ground truth. PLE is therefore highest at count 4 and lowest at count 1.

\begin{figure*}[!t]
    \centering
    \includegraphics[width=\textwidth]{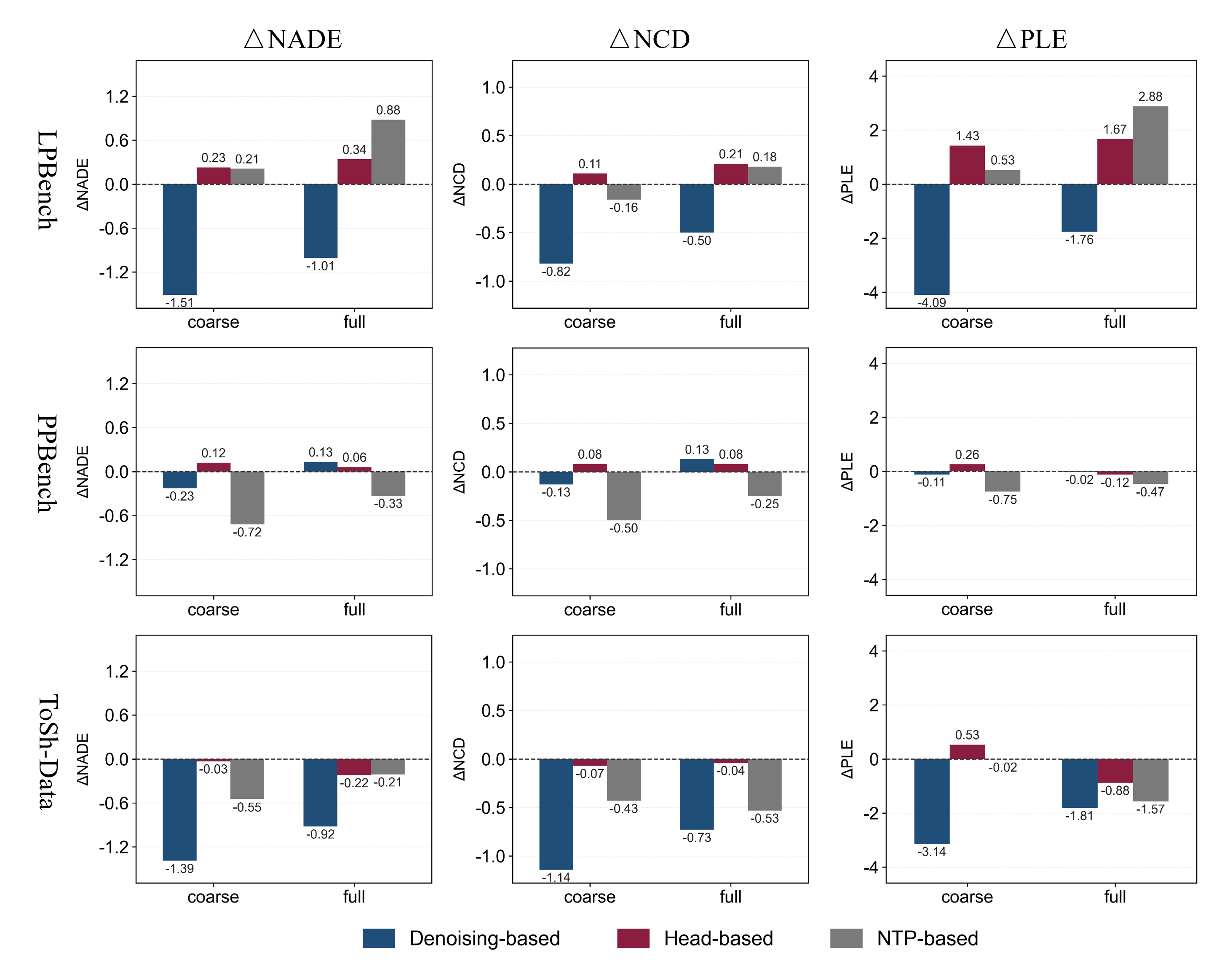}
    \caption{Ablation of \CoarseFull{} samples in pre-training across three methods and three benchmarks. \textbf{Each bar is the metric with \CoarseFull{} samples minus the metric without \CoarseFull{} samples.} Negative values mean lower metrics, hence better performance, with \CoarseFull{} samples.}\label{fig:7}
\end{figure*}

\subsection{Ablation of \CoarseFull{} Samples in Pre-training}
We compare pre-training with and without \CoarseFull{} samples, as shown in Fig.~\ref{fig:7}. For each architecture, removing these samples leaves the number of update steps unchanged. Path-related and VRSBench VQA samples are still loaded at a 3:1 ratio. This tests different data mixtures under the same update steps.

For the denoising-based method, including \CoarseFull{} samples reduces 16 of the 18 metrics across both inputs and all three benchmarks. This indicates that its late-stage pre-training bottleneck is mainly the VLM supplying denoising conditions rather than the path denoising expert itself. \CoarseFull{} samples strengthen the VLM and consequently reduce metrics. For the head-based method, metrics on LPBench and PPBench are generally higher with \CoarseFull{} samples than without them. Since this method directly learns path coordinates, spending more of a fixed update budget on paths themselves may be more effective. Its output head also directly decodes VLM features, and extra task-specific VLM training may create a mismatch between the head and VLM backbone and cause feature deviations.

For the NTP-based method, including \CoarseFull{} samples increases metrics on LPBench but reduces them on the other two benchmarks. This is related to the \CoarseFull{} task and benchmark characteristics. The \CoarseFull{} task requires the model to output image classes and spatial relationships around a path, incorporating reasoning from local to global. LPBench paths are short, so this reasoning may be an unnecessary additional capability for their generation. With fixed update counts, more direct path learning may therefore be more effective. Finally, all three methods generally have lower ToSh-Data metrics with \CoarseFull{} samples. ToSh-Data paths are mostly long, and its images cover wider real-world areas and contain more complex semantics than PPBench. Strengthening local-to-global reasoning in the VLM is therefore more helpful on this dataset.

Across the three methods and three benchmarks, 35 of 54 metrics decrease when \CoarseFull{} samples are included. Thus, under the current settings, adding \CoarseFull{} samples to the pre-training mixture is beneficial in a majority of cases. Its specific effect still depends on the architecture and benchmark characteristics, as we analyze above.

\begin{figure}[!ht]
    \centering
    \includegraphics[width=\columnwidth]{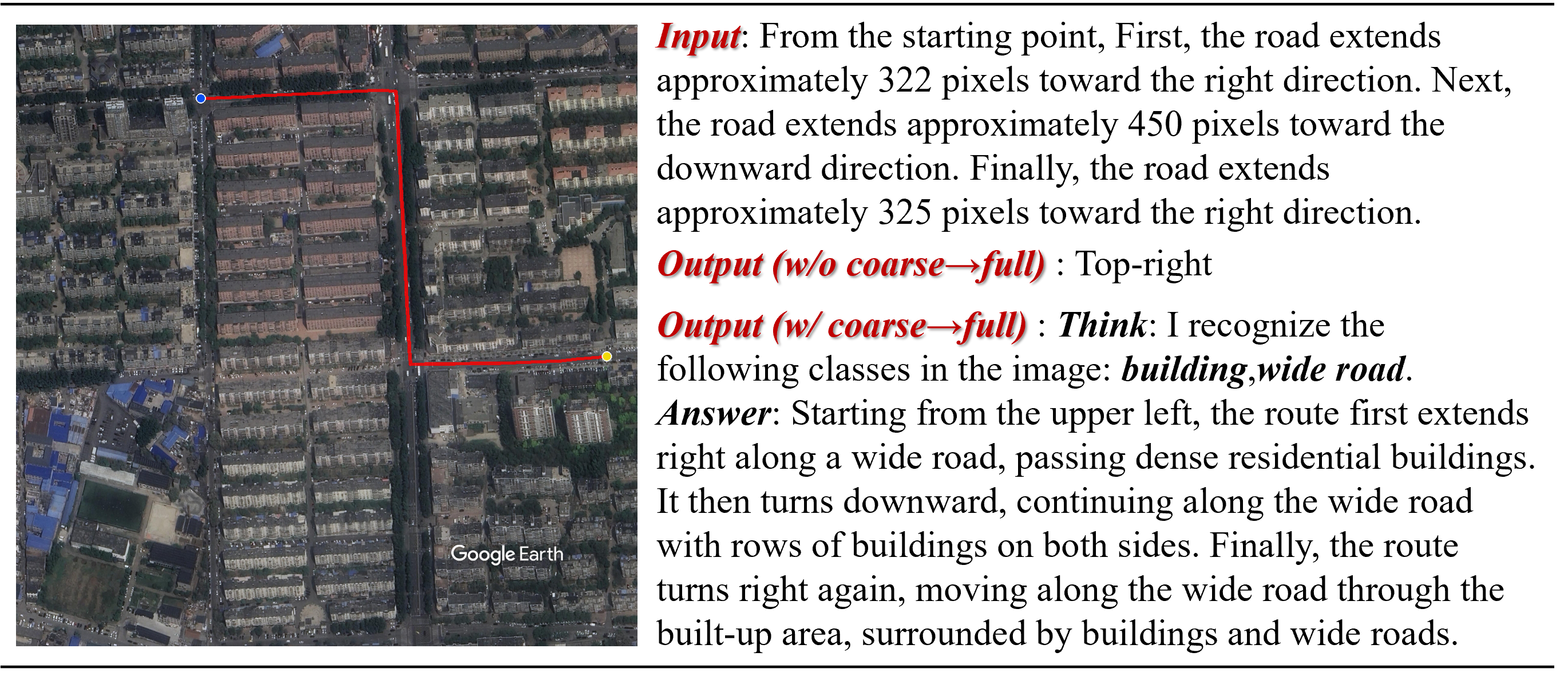}
    \caption{VLM outputs for \Coarse{} inputs after pre-training with and without \CoarseFull{} samples. Including \CoarseFull{} samples improves scene understanding.}\label{fig:8}
\end{figure}

\begin{figure*}[!t]
    \centering
    \includegraphics[width=\textwidth]{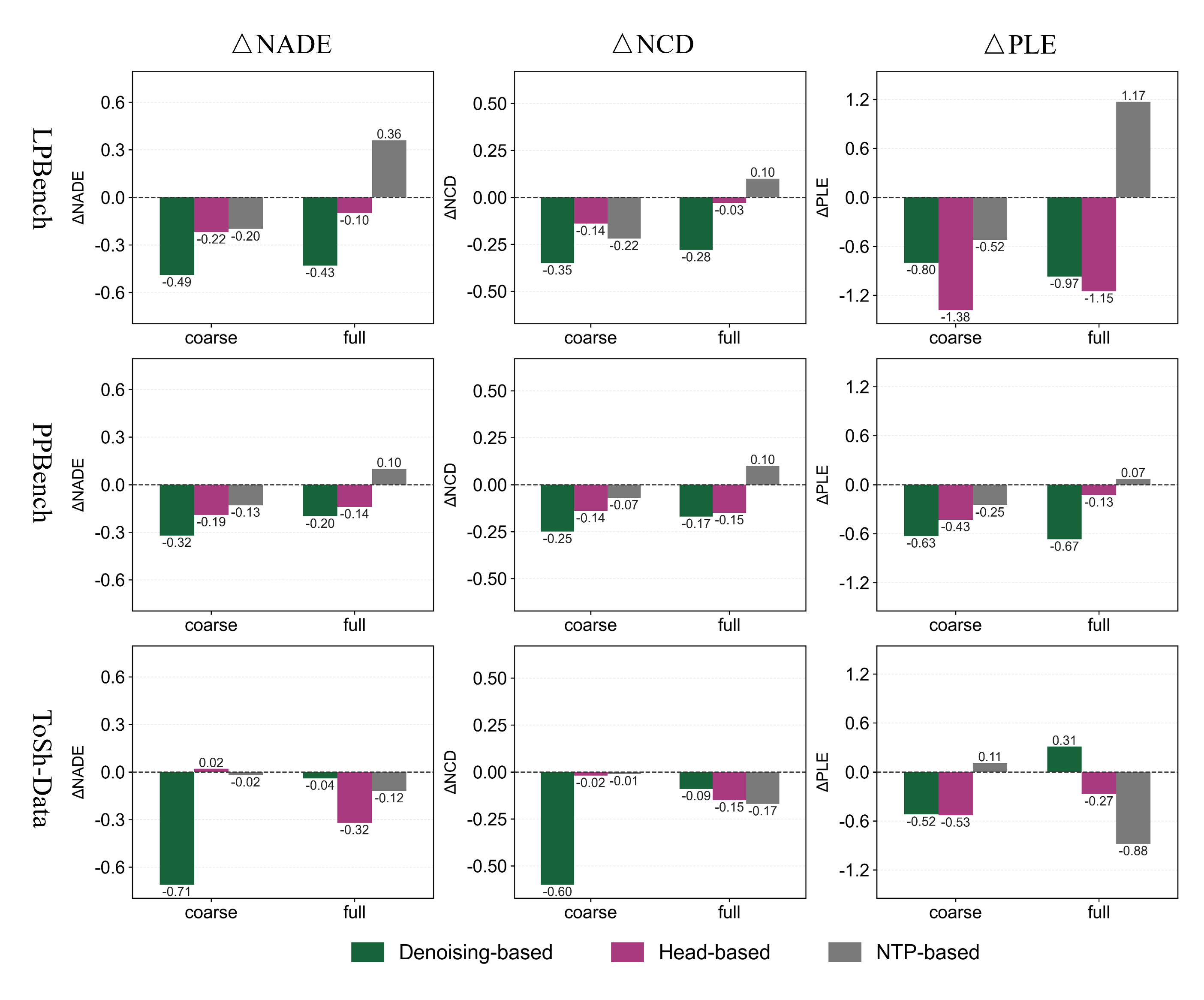}
    \caption{Ablation of the pre-training loss weight for \Coarse{} inputs across three methods and three benchmarks. \textbf{Each bar is the metric at weight 1.0 minus the metric at weight 0.6.} Negative values mean lower metrics, and hence better performance, at weight 1.0.}\label{fig:9}
\end{figure*}

We also visualize textual VLM outputs on one sample from ToSh-Data after pre-training with and without \CoarseFull{} samples, given \Coarse{}, an image, and start and end coordinates, as shown in Fig.~\ref{fig:8}. Including \CoarseFull{} samples greatly improves the model's ability to infer global path-related semantics from \Coarse{}. Without \CoarseFull{} samples, it still outputs the correct approximate path location in the image, indicating that the simple VQA samples from VRSBench provide basic language reasoning ability.

\begin{table*}[t]
\centering\caption{Head-based and Denoising-based performance after direct fine-tuning without RemotePath-175k task-specific pre-training, w/o GK, w/ GK-v2, or w/ GK-v1.}\label{tab:5}
\footnotesize\setlength{\tabcolsep}{1pt}\renewcommand{\arraystretch}{1.13}
\begin{tabular*}{\textwidth}{@{\extracolsep{\fill}}>{\raggedright\arraybackslash}m{2.25cm} l l rrrrrrrrrrrr@{}}
\toprule
\multirow{3}{*}{Model} & \multirow{3}{*}{GK setting} & \multirow{3}{*}{\shortstack[l]{Training steps\\(LPBench/PPBench)}} & \multicolumn{6}{c}{\Coarse{} input} & \multicolumn{6}{c}{\FullDesc{} input} \\
\cmidrule(lr){4-9}\cmidrule(lr){10-15}
 & & & \multicolumn{3}{c}{LPBench} & \multicolumn{3}{c}{PPBench} & \multicolumn{3}{c}{LPBench} & \multicolumn{3}{c}{PPBench} \\
\cmidrule(lr){4-6}\cmidrule(lr){7-9}\cmidrule(lr){10-12}\cmidrule(lr){13-15}
 & & & NADE & NCD & PLE & NADE & NCD & PLE & NADE & NCD & PLE & NADE & NCD & PLE \\
\midrule
\multirow{3}{2.25cm}{Head-based (ours)} & w/o GK & \auditcell{5!C4}{9000/6750} & \auditcell{5!D4}{4.64} & \auditcell{5!E4}{2.92} & \auditcell{5!F4}{28.55} & \auditcell{5!J4}{7.42} & \auditcell{5!K4}{5.28} & \auditcell{5!L4}{16.35}  & \auditcell{5!G4}{5.54} & \auditcell{5!H4}{3.56} & \auditcell{5!I4}{30.76} & \auditcell{5!M4}{8.03} & \auditcell{5!N4}{5.61} & \auditcell{5!O4}{17.27}  \\
 & w/ GK-v2 & \auditcell{5!C5}{9000/6750} & \auditcell{5!D5}{4.35} & \auditcell{5!E5}{2.72} & \auditcell{5!F5}{26.80} & \auditcell{5!J5}{7.05} & \auditcell{5!K5}{5.14} & \auditcell{5!L5}{13.62}  & \auditcell{5!G5}{5.32} & \auditcell{5!H5}{3.52} & \auditcell{5!I5}{27.12} & \auditcell{5!M5}{7.82} & \auditcell{5!N5}{5.58} & \auditcell{5!O5}{14.47}  \\
 & w/ GK-v1 & \auditcell{5!C6}{15,750/13,500} & \auditcell{5!D6}{4.14} & \auditcell{5!E6}{2.76} & \auditcell{5!F6}{21.51} & \auditcell{5!J6}{6.73} & \auditcell{5!K6}{4.99} & \auditcell{5!L6}{11.51}  & \auditcell{5!G6}{5.08} & \auditcell{5!H6}{3.51} & \auditcell{5!I6}{23.00} & \auditcell{5!M6}{7.43} & \auditcell{5!N6}{5.46} & \auditcell{5!O6}{11.57}  \\
\midrule
\multirow{3}{2.25cm}{Denoising-based (ours)} & w/o GK & \auditcell{5!C7}{9000/6750} & \auditcell{5!D7}{14.40} & \auditcell{5!E7}{6.25} & \auditcell{5!F7}{79.36} & \auditcell{5!J7}{21.78} & \auditcell{5!K7}{7.83} & \auditcell{5!L7}{80.26}  & \auditcell{5!G7}{14.30} & \auditcell{5!H7}{6.12} & \auditcell{5!I7}{79.33} & \auditcell{5!M7}{21.67} & \auditcell{5!N7}{7.79} & \auditcell{5!O7}{80.53}  \\
 & w/ GK-v2 & \auditcell{5!C8}{9000/6750} & \auditcell{5!D8}{17.49} & \auditcell{5!E8}{7.60} & \auditcell{5!F8}{82.83} & \auditcell{5!J8}{23.02} & \auditcell{5!K8}{8.14} & \auditcell{5!L8}{81.29}  & \auditcell{5!G8}{16.89} & \auditcell{5!H8}{7.34} & \auditcell{5!I8}{81.39} & \auditcell{5!M8}{22.92} & \auditcell{5!N8}{8.23} & \auditcell{5!O8}{81.41}  \\
 & w/ GK-v1 & \auditcell{5!C9}{15,750/13,500} & \auditcell{5!D9}{12.98} & \auditcell{5!E9}{6.03} & \auditcell{5!F9}{75.32} & \auditcell{5!J9}{23.37} & \auditcell{5!K9}{8.44} & \auditcell{5!L9}{80.18}  & \auditcell{5!G9}{12.54} & \auditcell{5!H9}{5.76} & \auditcell{5!I9}{74.76} & \auditcell{5!M9}{23.15} & \auditcell{5!N9}{8.35} & \auditcell{5!O9}{80.23}  \\
\bottomrule
\end{tabular*}
\end{table*}

\begin{table*}[t]
\centering\caption{Numbers of format-valid predictions of the NTP-based method after direct fine-tuning without RemotePath-175k task pre-training, w/o GK, w/ GK-v2, or w/ GK-v1.}\label{tab:6}
\small\setlength{\tabcolsep}{3pt}
\begin{tabular}{lllcccc}
\toprule
\multirow{2}{*}{Model} & \multirow{2}{*}{GK setting} & \multirow{2}{*}{\shortstack[l]{Training steps\\(LPBench/PPBench)}} &  \multicolumn{2}{c}{LPBench} & \multicolumn{2}{c}{PPBench} \\
 &  &  & \Coarse{} & \FullDesc{} & \Coarse{} & \FullDesc{} \\
\midrule
\multirow{3}{*}{NTP-based} & w/o GK & \auditcell{6!C3}{9000/6750} & \auditcell{6!D3}{227/2047(11.1\%)} & \auditcell{6!E3}{211/2047(10.3\%)} & \auditcell{6!F3}{58/1483(3.9\%)} & \auditcell{6!G3}{40/1483(2.7\%)} \\
 & w/ GK-v2 & \auditcell{6!C4}{9000/6750} & \auditcell{6!D4}{181/2047(8.8\%)} & \auditcell{6!E4}{211/2047(10.3\%)} & \auditcell{6!F4}{42/1483(2.8\%)} & \auditcell{6!G4}{51/1483(3.4\%)} \\
 & w/ GK-v1 & \auditcell{6!C5}{15,750/13,500} & \auditcell{6!D5}{324/2047(15.8\%)} & \auditcell{6!E5}{281/2047(13.7\%)} & \auditcell{6!F5}{69/1483(4.7\%)} & \auditcell{6!G5}{74/1483(5.0\%)} \\
\bottomrule\end{tabular}
\end{table*}

\subsection{Why \Coarse{} input Often Gives Lower Metrics Than \FullDesc{} input?}
Table~\ref{tab:3} shows generally lower metrics for \Coarse{} than for \FullDesc{} across our three architectures and benchmarks. We consider whether the \Coarse{} pattern is easier to learn, leaving \FullDesc{} insufficiently learned. To investigate this, we reduce the pre-training loss weight of \Coarse{} samples to 0.6 while keeping \FullDesc{} at 1.0 and retaining \CoarseFull{} samples. By reducing learning from the more direct \Coarse{} input, we examine the effect on zero-shot performance, as shown in Fig.~\ref{fig:9}.

For head-based and denoising-based methods, reducing the \Coarse{} weight to 0.6 generally increases metrics on all three benchmarks. Greater reliance on \FullDesc{} therefore does not provide more effective training. The explicit directions and distances in \Coarse{} help the model learn path patterns without adding redundant global information. This further suggests that the generally lower \Coarse{} metrics do not result from insufficient \FullDesc{} learning or competition between the two inputs in the loss. Rather, redundancy in \FullDesc{} makes the most useful path information difficult to extract when relying on these samples. At test time, the same difficulty can explain the lower \Coarse{} metrics. For NTP-based generation, reducing the \Coarse{} pre-training weight lowers \FullDesc{} metrics on LPBench and PPBench, consistent with the larger relative \FullDesc{} loss weight. Even so, \Coarse{} metrics on both benchmarks remain lower than \FullDesc{} metrics.

\subsection{Comparison With and Without Task-Specific Pre-training}
We evaluate direct fine-tuning of all three architectures on LPBench and PPBench without RemotePath-175k task-specific pre-training, including settings w/ GK-v2, w/ GK-v1 and w/o GK. We follow the update counts and loading ratios for samples specified above for SFT without task-specific pre-training. For each setting, including w/o GK, w/ GK-v1, and w/ GK-v2, we report results after three times the update count of one complete traversal, to promote convergence.

Table~\ref{tab:5} reports metrics only for head-based and denoising-based methods. The NTP-based method produces very few format-valid predictions under every setting without task-specific pre-training, and metrics calculated on these valid predictions are much higher than those of the other architectures. Its valid counts are reported in Table~\ref{tab:6}. This suggests that NTP-based generation generally needs large-scale task-specific pre-training to realize its potential and may perform poorly without it.

For the head-based method, after three times the update count, most metrics on both benchmarks are lowest w/ GK-v1, second-lowest w/ GK-v2, and highest w/o GK. Additional general knowledge therefore improves its performance without task-specific pre-training, and more general knowledge gives larger gains. Denoising-based metrics are generally much higher than head-based metrics on both benchmarks, and the same improvement with added GK is not observed. This indicates inadequate convergence in this setting and suggests that the denoising-based method requires more training data than the head-based method to perform well.

Even the best values in Table~\ref{tab:5} remain higher than the zero-shot metrics after task-specific pre-training on LPBench and PPBench at most cases. These results indicate that relatively large-scale, diverse task-specific pre-training is necessary to obtain a task-specific VLM for path generation on remote sensing images.

\section{Limitations}

Our experiments show that pre-training on the constructed dataset allows all three methods to converge reasonably well and develop zero-shot capability. They outperform some commercial VLMs on some benchmarks, but remain behind advanced models such as Gemini 3.6 Flash in other cases. Commercial models accumulate data and iterate quickly, allowing continued improvement through long-term experience. Our pre-training dataset may also remain too small to fully realize the potential of methods such as NTP-based generation. However, we document a scalable construction strategy, especially the use of generative models and road skeleton extraction to correctly align images and paths. This strategy can produce larger and more diverse datasets. Our experiments support both the effectiveness of the current pre-training dataset and the construction strategy. Larger and more diverse datasets built with this strategy should further improve the zero-shot performance of the three architectures.

\section{Conclusion}

We investigate task-specific VLMs for path generation from top-down remote sensing images. We explore three VLM architectures and use a scalable data construction method to build a relatively large and diverse pre-training dataset. Pre-training enables all three methods to achieve zero-shot performance that exceeds some advanced commercial models in some settings. We also construct three benchmarks supporting stratified evaluation and three metrics for quantitative evaluation. Through extensive experiments, we explore this problem and hope to encourage further research.

\section*{Acknowledgements}

This work was supported by Intelligent Computing Platform of BIT.

\bibliographystyle{IEEEtran}
\bibliography{remotevlpg}
\end{document}